%% file: main.tex
\documentclass[10pt,twocolumn,letterpaper]{article}

\usepackage{cvpr}              
\input{preamble}
\definecolor{cvprblue}{rgb}{0.21,0.49,0.74}
\usepackage[pagebackref,breaklinks,colorlinks,allcolors=cvprblue]{hyperref}
\usepackage{textcomp}
\usepackage{algorithm}
\usepackage{algpseudocode}
\usepackage{graphicx} 
\usepackage{multirow}
\usepackage{makecell}

\usepackage{graphicx}
\usepackage{amsmath}
\usepackage{amssymb}
\usepackage{booktabs}
\usepackage{svg}

\newenvironment{acks}{\section*{Acknowledgments}}{}
\usepackage[accsupp]{axessibility}  

\def\paperID{8459} 
\def\confName{CVPR}
\def\confYear{2026}

\title{SEA-Vision: A Multilingual Benchmark for Comprehensive Document and Scene Text Understanding in Southeast Asia}

\author{
Pengfei Yue$^{1}$\thanks{Equal contribution.} \and
Xingran Zhao$^{2}$\footnotemark[1] \and
Juntao Chen$^{3}$\footnotemark[1] \and
Peng Hou$^{2}$ \and
Wang Longchao$^{2}$ \and
Jianghang Lin$^{1}$ \and
Shengchuan Zhang$^{1}$\thanks{Corresponding author} \and
Anxiang Zeng$^{2}$ \and
Liujuan Cao$^{1}$ \\
$^{1}$Xiamen University, China\
\hspace{1cm}$^{2}$Shopee, China\
\hspace{1cm}$^{3}$Tongji University, China\
}

\begin{document}
\maketitle

\input{sec/0_abstract}

\input{sec/1_intro}
\input{sec/2_related_work}
\input{sec/3_main}
\input{sec/4_dataset_statistics}
\input{sec/5_evaluation}
\input{sec/6_benchmark}

\input{sec/7_conclusion}
\begin{acks}

This work was supported by  National Key Research and Development Program of China (No. 2025YFE0113500), National Science Fund for Distinguished Young Scholars (No.62025603 and No.62525605), National Natural Science Foundation of China (No. U21B2037, U22B2051, No. U23A20383, No. 62176222, No. 62176226, No. 62272401, No. 62576300).

\end{acks}

{
    \small \bibliographystyle{ieeenat_fullname}
    \bibliography{main_ref}
}


\clearpage
\appendix
\renewcommand{\thesection}{\Alph{section}}

\setcounter{figure}{0}
\setcounter{table}{0}
\setcounter{equation}{0}
\renewcommand{\thefigure}{A\arabic{figure}}
\renewcommand{\thetable}{A\arabic{table}}
\renewcommand{\theequation}{A\arabic{equation}}

\twocolumn[
\begin{center}
    \Large\bfseries SEA-Vision: A Multilingual Benchmark for Comprehensive Document and Scene Text Understanding in Southeast Asia\par
    \vspace{0.5em}
    \large Appendix\par
    \vspace{1em}
\end{center}
]

\input{Appendix/Appendix}


\end{document}

%% file: sec/0_abstract.tex
\begin{abstract}
Multilingual document and scene text understanding plays an important role in applications such as search, finance, and public services. 
However, most existing benchmarks focus on high-resource languages and fail to evaluate models in realistic multilingual environments. 
In Southeast Asia, the diversity of languages, complex writing systems, and highly varied document types make this challenge even greater. 
We introduce \textbf{SEA-Vision}, a benchmark that jointly evaluates Document Parsing and Text-Centric Visual Question Answering~(TEC-VQA)  across 11 Southeast Asian languages.
SEA-Vision contains 15,234 document parsing pages from nine representative document types, annotated with hierarchical page-, block-, and line-level labels.
It also provides 7,496 TEC-VQA question–answer pairs that probe text recognition, numerical calculation, comparative analysis, logical reasoning, and spatial understanding.
To make such multilingual, multi-task annotation feasible, we design a hybrid pipeline for Document Parsing and TEC-VQA.
It combines automated filtering and scoring with MLLM-assisted labeling and lightweight native-speaker verification, greatly reducing manual labeling while maintaining high quality.
We evaluate several leading multimodal models and observe pronounced performance degradation on low-resource Southeast Asian languages, highlighting substantial remaining gaps in multilingual document and scene text understanding. 
We believe SEA-Vision will help drive global progress in document and scene text understanding.
\end{abstract}

%% file: sec/1_intro.tex
\begin{figure}[t!] 
\centering 
\includegraphics[width=0.48\textwidth]{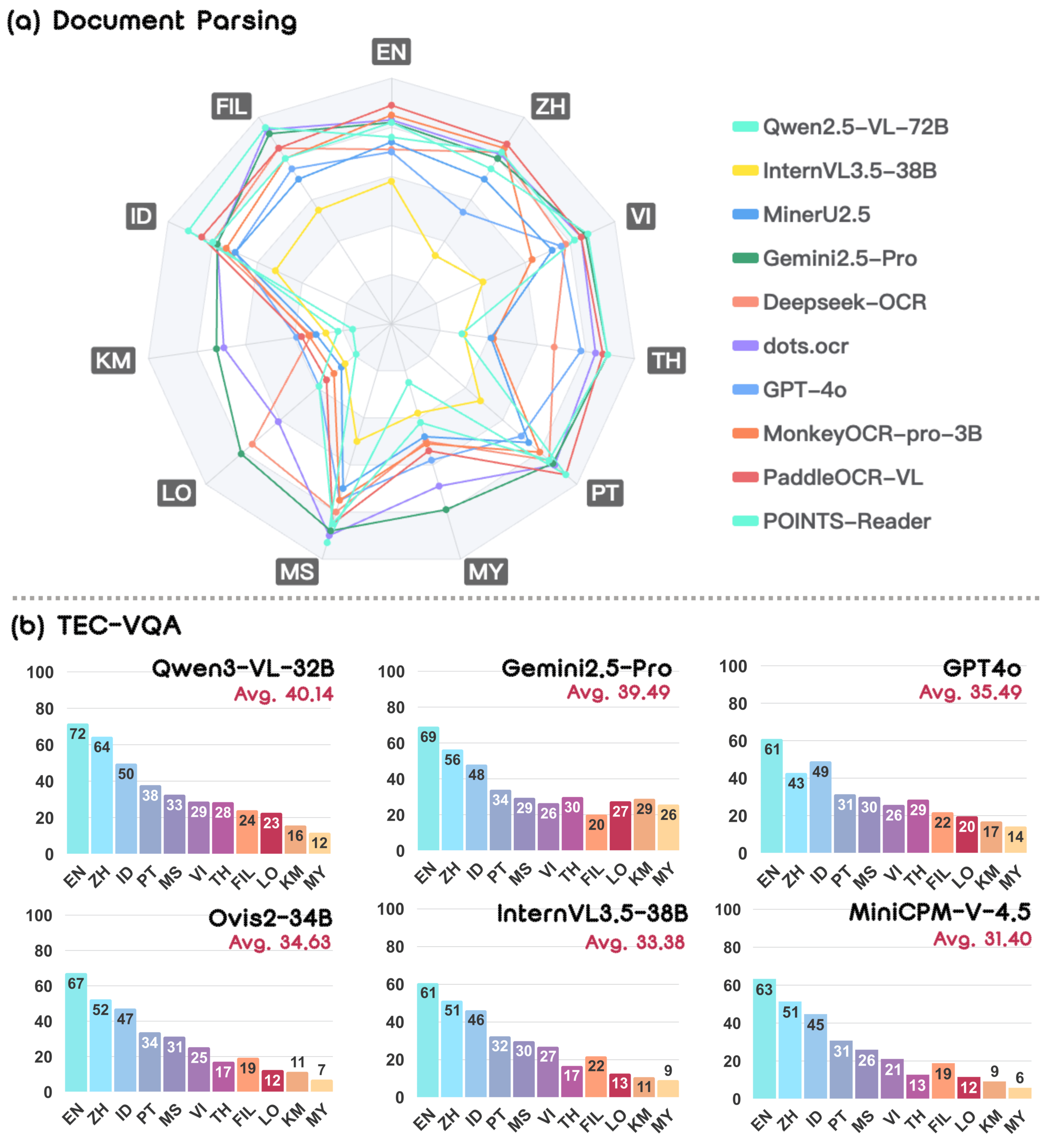} 
\caption{\textbf{Performance of representative models on SEA-Vision.} (a) End-to-end text recognition performance for document parsing across 11 languages. (b) TEC-VQA accuracy by language and model, along with overall averages~(Avg.).}
\vspace{-8px}
\label{fig: motivation} 
\end{figure}

\section{Introduction}
\label{sec:intro}
With the rapid deployment of Multimodal Large Language Models~(MLLMs) \cite{hurst2024gpt, qwen3technicalreport, wang2025internvl3_5} in search, finance, government, and education, document and scene text understanding \cite{ouyang2025omnidocbench, liu2024ocrbench, fu2024ocrbenchv2} has become a core capability for real-world applications. In text-rich images, textual content often carries the central semantics, making accurate text perception and interpretation essential for text-centric visual understanding. Within this scope, Document Parsing \cite{liu2024focus, ouyang2025omnidocbench} targets the extraction of structured, machine-readable content from unstructured documents, while Text-Centric Visual Question Answering (TEC\mbox{-}VQA) \cite{mathew2021docvqa,mathew2022infographicvqa,singh2019textvqa} emphasizes reasoning over textual information across diverse scenes to answer complex questions.

\begin{figure*}[!htbp] 
\centering 
\includegraphics[width=\textwidth]{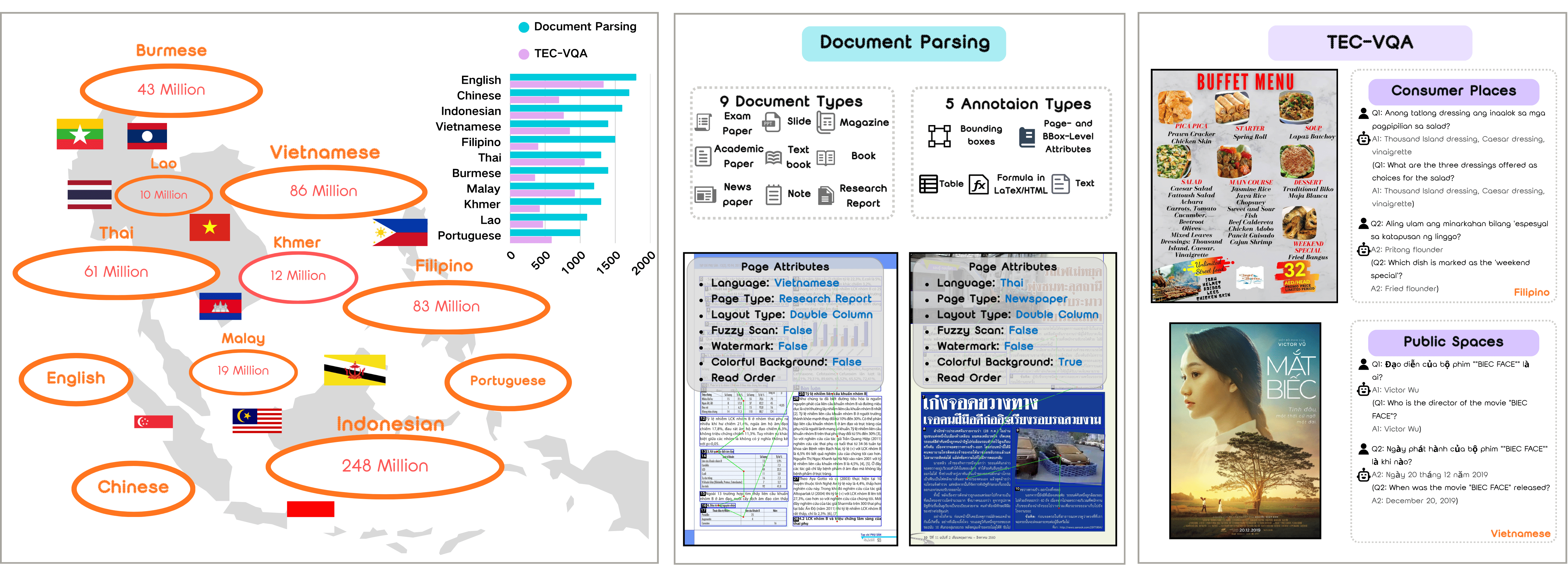} 
\caption{\textbf{SEA-Vision benchmark overview.} Geographical language coverage and dataset scale (left), Document Parsing types and sample page attributes (middle), and TEC-VQA examples across consumer places and public spaces (right).}
\label{fig: overview} 
\end{figure*}

Despite progress, existing benchmarks fall short of realistic multilingual conditions. 
Most datasets \cite{biten2019scenetextvqa, singh2019textvqa, fu2024ocrbenchv2, liu2024ocrbench} center on English or other high-resource languages, offering limited evidence of generalization to low-resource scripts. 
Many document-focused datasets \cite{li2020docbank, pfitzmann2022doclaynet} concentrate on narrow domains such as academic papers and overlook diverse real-world formats such as exams, notes, newspapers, and product images. 
Moreover, Document Parsing and TEC\mbox{-}VQA are often evaluated independently, preventing unified assessment across recognition, structure, and high-level reasoning. 
In multilingual settings, Optical Character Recognition~(OCR) or translation-based annotation strategies \cite{jiu2025tvqacml} further introduce visual–semantic misalignment.
These gaps are amplified in \textbf{Southeast Asia (SEA)}, a linguistically rich region spanning Latin, Brahmic, Arabic-based, and logographic scripts. 
As shown in \cref{fig: overview}, complex and diverse multilingual content, dense layouts, and heterogeneous visual structures are common in real applications.
Yet no existing benchmark jointly covers major SEA languages while enabling cross-task, cross-script evaluation, limiting progress in building robust multilingual systems.
To address this, we present \textbf{SEA-Vision}, a comprehensive benchmark for multilingual document and scene text understanding across 11 Southeast Asian languages—\textit{English (EN), Chinese (ZH), Vietnamese (VI), Thai (TH), Filipino (FIL), Malay (MS), Indonesian (ID), Lao (LO), Khmer (KM), Burmese (MY), and Portuguese (PT)}. 
SEA-Vision consists of \textbf{15{,}234} document parsing samples covering nine representative document types, annotated with hierarchical page-, block-, and line-level labels. 
It further includes \textbf{7{,}496} TEC\mbox{-}VQA question–answer pairs spanning five key reasoning skills: text recognition, numerical calculation, comparative analysis, logical reasoning, and spatial understanding.
To make multilingual, multi-task annotation feasible at scale, we designed a hybrid pipeline that integrates automatic processing, MLLM-assisted labeling, and native-speaker verification, significantly reducing manual annotation costs and improving overall efficiency. For \emph{TEC\mbox{-}VQA}, we tackle visual-textual misalignment by re-rendering translated text back into the image and using MLLMs to generate multilingual question-answer candidates, which are then filtered through cross-language consistency checks and native-speaker review to ensure alignment with the visible content.
Finally, we establish a unified evaluation framework across both tasks. For Document Parsing, we benchmark three main paradigms—\emph{pipeline models}, \emph{expert models}, and \emph{general models}—in an end-to-end setting.
For TEC\mbox{-}VQA, we evaluate leading open-source and proprietary MLLMs under a consistent zero-shot protocol. As summarized in Fig.~\ref{fig: motivation}, models that perform well on high-resource languages such as English and Chinese degrade substantially on many SEA languages. In summary, our benchmark introduces the following key contributions:
\begin{itemize}
    \item \textbf{Unified benchmark and coverage.} SEA-Vision jointly evaluates \emph{Document Parsing} and \emph{TEC\mbox{-}VQA} under a single framework, spanning \textbf{11 SEA languages} and multiple scripts with data from real-world sources.
    \item \textbf{High-quality data and process.} We release \textbf{15{,}234} document samples and \textbf{7{,}496} scene text question--answer pairs constructed via automated pipelines and native-speaker verification to ensure visual--semantic alignment and consistent annotations.
    \item \textbf{Baselines and empirical insights.} Benchmarking shows 3–5$\times$ higher NED \cite{lcvenshtcin1966binary} for Document Parsing and 5–7$\times$ lower accuracy for TEC\mbox{-}VQA on low-resource SEA scripts—highlighting the need for improved multilingual document and scene text understanding in MLLMs.
\end{itemize}

%% file: sec/2_related_work.tex
\section{Related Work}
\label{sec:related_work}

\subsection{Document Parsing}
Existing approaches to document parsing can be broadly grouped into three categories. (1) \emph{Pipeline Models} \cite{cui2025paddleocr, feng2025dolphin, li2025monkeyocr, wang2024mineru, niu2025mineru2} decompose parsing into modular stages such as document layout detection \cite{zhao2024doclayout, gu2021unidoc, huang2022layoutlmv3, pramanik2020towards}, optical character recognition \cite{huang2022swintextspotter, li2022ppocrv3, liu2020abcnet, smith2009adapting, wang2021pgnet}, formula recognition \cite{li2020improvingformula, wang2024unimernet, zhang2018multiformula}, and table recognition \cite{huang2020tabtransformer, huang2023improvingtable, li2022ppocrv3}, and then fuse the results into structured outputs (e.g., Markdown, JSON, HTML). Representative systems like Marker\footnote{\url{https://github.com/datalab-to/marker}} and MinerU \cite{wang2024mineru} integrate off-the-shelf detectors and recognizers, optionally augmented with expert models to refine reading order \cite{wang2021layoutreader, gu2022xylayoutlm}, merge cross-page tables, and repair inline math, achieving efficient parsing at the cost of error accumulation across stages. (2) \emph{General Models} such as GPT-4o \cite{hurst2024gpt} and Qwen-VL-Series \cite{Qwen-VL,Qwen2-VL,Qwen2.5-VL,qwen3technicalreport} treat document parsing as a single sequence generation problem, directly producing structured content from document images after large-scale pre-training on OCR-style corpora, and showing strong zero-shot performance without task-specific fine-tuning. (3) \emph{Expert Models} \cite{dots.ocr,liu2025points,wei2025deepseekocr,ye2023mplug} further tailor VLM-style architectures to document understanding, jointly modeling layout, reading order, tables, and formulas under a unified objective to better balance accuracy and efficiency than both pure pipeline systems and general-purpose MLLMs. However, nearly all of these methods are developed and evaluated on high-resource languages, primarily English and Chinese, leaving their behavior on low-resource language pages largely untested.

\subsection{Text-Centric Visual Question Answering}

Text-centric Visual Question Answering (TEC-VQA) \cite{mathew2021docvqa,mathew2022infographicvqa,singh2019textvqa} builds on advances in Visual Question Answering (VQA) \cite{antol2015vqa} enabled by instruction-tuned MLLMs. When trained on visual text understanding corpora, these models exhibit strong zero-shot performance on benchmarks such as MTVQA \cite{tang2025mtvqa}. A series of works extend LLaVA-style models to document domains (e.g., LLaVAR \cite{zhang2023llavar}, UniDoc \cite{feng2023unidoc}) by training them to predict text and layout coordinates directly from document images, while others—such as DocPedia \cite{feng2024docpedia}—operate in the frequency domain to support higher-resolution inputs without increasing sequence length. Models including mPLUG-DocOwl \cite{ye2023mplug}, Qwen-VL-Series \cite{Qwen-VL,Qwen2-VL,Qwen2.5-VL,qwen3technicalreport}, TextMonkey \cite{liu2024textmonkey}, and TextHarmony \cite{zhao2024harmonizing} further leverage document-focused VQA datasets and unify visual text generation and comprehension within a single architecture, thereby improving reasoning over numbers, layouts, and complex text-centric scenes. Despite these advances, existing datasets and models remain heavily skewed toward high-resource languages such as English and Chinese, offering limited coverage of 

\input{Tables/table4_relatedwork}

\noindent low-resource languages and mixed-script scenarios common in Southeast Asia. This gap motivates the need for a multilingual benchmark that jointly evaluates document parsing and TEC-VQA capabilities.

\subsection{Document and Scene Text Understanding Benchmark}
A rich line of benchmarks has been proposed for document and scene text understanding, but they each cover only part of the space summarized in Tab.~\ref{table4: relatedwork}. 
Early scene text and OCR benchmarks such as ICDAR2013/2015 \cite{karatzas2013icdar,karatzas2015icdar} and MLT-2019 \cite{nayef2019icdar2019} mainly target detection and recognition, with a strong bias toward English or a few high-resource languages and without structured Document Parsing or reasoning tasks. 
Document-centric VQA benchmarks such as DocVQA \cite{mathew2021docvqa}, TextVQA \cite{singh2019textvqa}, and InfographicVQA \cite{mathew2022infographicvqa}, as well as multilingual extensions like M3Exam \cite{zhang2023m3exam} and MTVQA \cite{tang2025mtvqa}, introduce question answering over text-rich images and begin to explore multilingual settings. Most of these benchmarks construct question–answer pairs by leveraging OCR outputs or machine translation, which can introduce visual–text misalignment and rarely provide page-level structural annotations. In contrast, MTVQA is built from carefully human-authored multilingual question–answer pairs, avoiding OCR or translation-induced noise but at the cost of substantial manual annotation effort.
More recent document parsing and OCR evaluation suites such as Fox \cite{liu2024focus}, OmniDocBench \cite{ouyang2025omnidocbench} and OCRBench v2 \cite{fu2024ocrbenchv2} move toward fine-grained layout, table, and formula understanding, but are largely restricted to English and Chinese. 
In contrast, SEA-Vision is designed as a unified benchmark that jointly evaluates Document Parsing and TEC-VQA across 11 languages, including seven low-resource Southeast Asian languages, with carefully human-verified annotations to ensure high-fidelity alignment between visible text, structure, and semantic supervision.

%% file: Tables/table4_relatedwork.tex
\begin{table}[ht]
\centering
\large
\renewcommand{\arraystretch}{1.25} 
\resizebox{0.48\textwidth}{!}{
\begin{tabular}{l|ccccc}
\hline
Benchmark & Langs & Low-Resource & DP Images & \makecell{TEC-VQA\\QA Pairs} & 
\makecell{Unified Eval\\(DP+VQA)} \\
\hline
ICDAR2013/2015 \cite{karatzas2013icdar, karatzas2015icdar} & 1 & $\times$ & $\times$ & $\times$ & $\times$ \\
DocVQA \cite{mathew2021docvqa}        & 1 & $\times$ & $\times$ & 50{,}000 & $\times$ \\
TextVQA \cite{singh2019textvqa}        & 1 & $\times$ & $\times$ & 45{,}336 & $\times$ \\
MLT-2019 \cite{nayef2019icdar2019}       & 10 & $\checkmark$ (2) & $\times$ & $\times$ & $\times$ \\
M3Exam \cite{zhang2023m3exam}        & 9  & $\checkmark$ (6) & $\times$ & 12{,}317 & $\times$ \\
MTVQA \cite{tang2025mtvqa}         & 9  & $\checkmark$ (2) & $\times$ & 6{,}778  & $\times$ \\
Fox \cite{liu2024focus}           & 2 (EN/ZH) & $\times$ & 212   & $\times$ & $\times$ \\
OmniDocBench \cite{ouyang2025omnidocbench}  & 2 (EN/ZH) & $\times$ & 981   & $\times$ & $\times$ \\
OCRBench v2 \cite{fu2024ocrbenchv2}   & 2 (EN/ZH) & $\times$ & 2{,}400 & 7{,}600 & \checkmark \\
CC-OCR \cite{yang2025cc}        & 10 & $\checkmark$ (1) & 800  & 6{,}258 & \checkmark \\
\hline
SEA-Vision (Ours) & 11 & $\checkmark$ (7) & 15{,}234 & 7{,}496 & $\checkmark$ \\
\hline
\end{tabular}
}
\caption{Comparison of existing text-related benchmarks. ``Low-Resource'' denotes the number of low-resource languages included. ``Unified Eval'' indicates whether the benchmark jointly evaluates Document Parsing (DP, Document Parsing) and Text-Centric Visual Question Answering (TEC-VQA).}
\vspace{-10px}
\label{table4: relatedwork}
\end{table}

%% file: sec/3_main.tex
\section{SEA-Vision Benchmark} 

\begin{figure*}[!htbp] 
\centering 
\includegraphics[width=\textwidth]{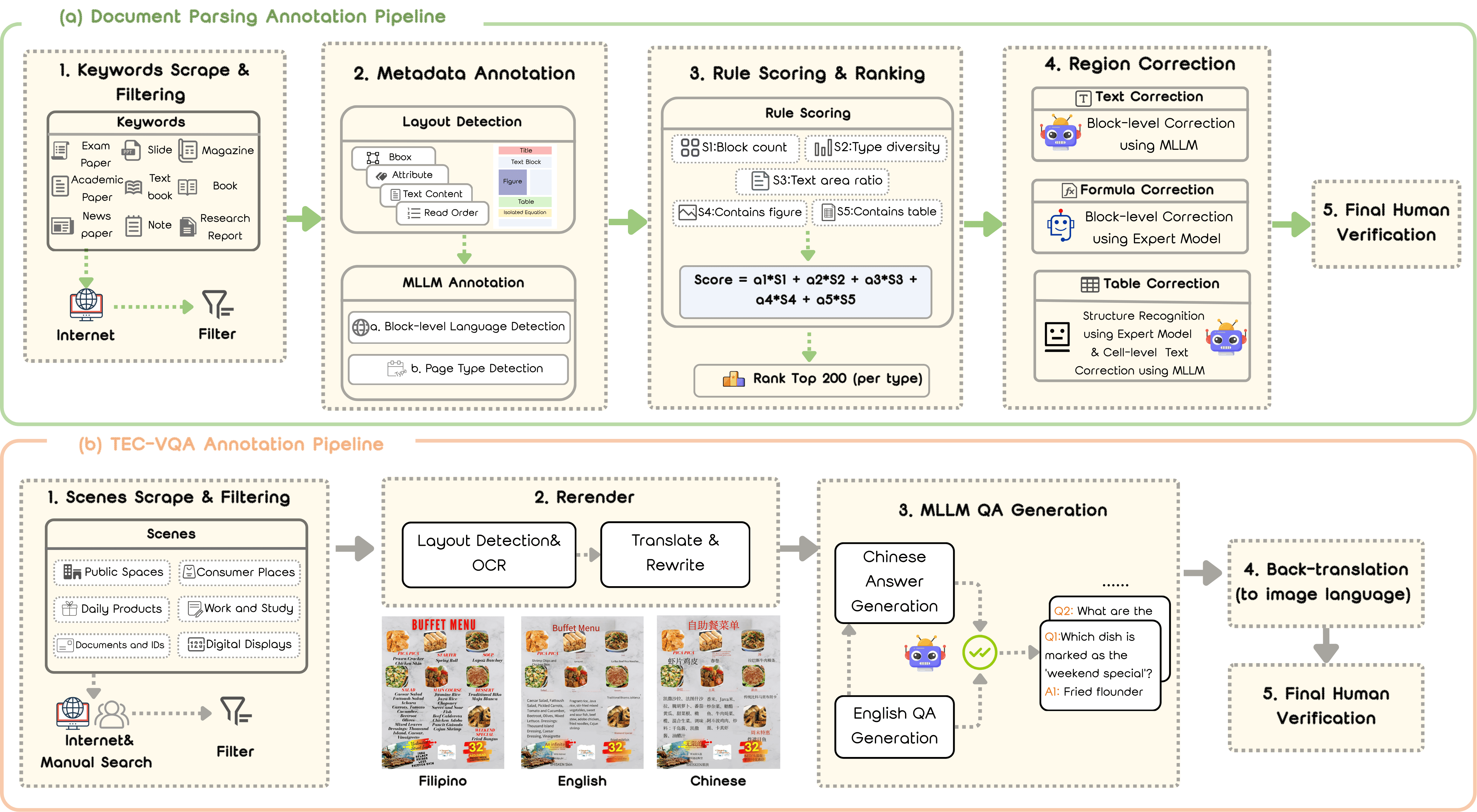} 
\caption{
\textbf{Overview of the data annotation pipelines.}
(a) \textit{Document Parsing Annotation Pipeline}: Internet-sourced document pages are first collected using domain-specific keywords and filtered for quality. Metadata annotation includes layout detection and MLLM–based analysis for language and page type identification. Candidate pages are ranked by a rule-based scoring function considering block count, type diversity, text area ratio, and presence of figures or tables. Selected samples undergo region-level correction via specialized models for text, formulas, and tables, followed by final human verification.
(b) \textit{TEC-VQA Annotation Pipeline}: Scene images from diverse environments (e.g., public spaces, consumer places, documents) are gathered and filtered. Layout and text are detected and re-rendered with multilingual content. An MLLM first generates English QA pairs; the English questions are then translated into Chinese to obtain Chinese QA, which is aligned with the English QA for consistency. The resulting bilingual QA pairs are translated into the image language and manually verified.
}
\vspace{-10px}
\label{fig: main} 
\end{figure*}

We introduce SEA-Vision, a benchmark of real-world documents and images designed to evaluate multilingual document and scene text understanding across diverse Southeast Asian languages.
It encompasses two related tasks: (1) Document Parsing, which evaluates structured content extraction from documents, and (2) TEC-VQA, which evaluates question answering from images with multilingual text. 
\subsection{Document Parsing}

\subsubsection{Data Acquisition}
To construct a large-scale and diverse document parsing dataset, we first perform targeted web scraping using domain-specific keywords covering various document types.
%
The keywords are designed to ensure coverage of multiple content categories and visual layouts. Retrieved results are automatically filtered to remove duplicates, non-document and low-resolution pages, resulting in a corpus of approximately 3M PDF documents for subsequent document parsing annotation.
%

\subsubsection{Data Annotation Pipeline}

As shown in Fig.~\ref{fig: main}, the document parsing annotation follows a four-stage pipeline integrating automatic detection, rule-based selection, multimodal correction, and human verification. 

\noindent\textbf{Metadata Annotation.}
For each retained page, a layout detection model segments the document into 10 predefined region categories, including title, text block, table, formula, figure, etc. 
Each region is assigned bounding box coordinates, textual content, and local reading order. 
A multimodal large language model then performs block-level language detection and page-type classification to enrich metadata. 
This metadata annotation serves as a coarse categorization step, ensuring that documents in low-resource Southeast Asian languages are preserved during subsequent filtering and not inadvertently removed due to their rarity or nonstandard layouts.

\begin{algorithm}[h]
\caption{Language–page-type aware filtering, scoring, and ranking}
\label{alg1}
\begin{algorithmic}[1]
\Require Pages $\mathcal{P}$; \texttt{ALLOWED\_LANGS}, \texttt{CAP\_DET}, \texttt{CAP\_TYPES}, \texttt{FIG\_TYPES}; optional $K$ or threshold $\tau$
\State $\mathcal{P}_{\mathrm{keep}}\gets \varnothing$
\ForAll{$p\in\mathcal{P}$}
  \State $L\gets \text{Split}(\text{lang\_raw}(p),\texttt{\_})$
  \If{$|\text{layout\_dets}(p)|<1$ \textbf{or} $(\exists d_i:\ \text{ignore}_i=\texttt{True})$} \textbf{continue} \EndIf
  \If{$|L|>2$ \textbf{or} $(L\not\subseteq \texttt{ALLOWED\_LANGS})$} \textbf{continue} \EndIf
  \State $N\gets |\text{layout\_dets}(p)|$
  \State $A_{\text{text}}\gets \sum_{d_i:\ \text{category\_type}_i\in \text{TEXT\_TYPES}}\!\text{Area}(\text{polygon}_i)$
  \State $T\gets A_{\text{text}}/\text{PageArea}(p)$
  \State $C\gets \big|\{\text{category\_type}_i\}_i\big|$
  \State $b_{\text{tab}}\gets \mathbf{1}\{\exists d_i:\ \text{category\_type}_i=\texttt{table}\}$
  \State $b_{\text{fig}}\gets \mathbf{1}\{\exists d_i:\ \text{category\_type}_i\in \texttt{FIG\_TYPES}\}$
  \State $\text{Score}(p)\gets 30\cdot \min(1,N/\texttt{CAP\_DET})+30\cdot \mathrm{clip}(T,0,1)+20\cdot \min(1,C/\texttt{CAP\_TYPES})+10\cdot b_{\text{tab}}+10\cdot b_{\text{fig}}$
  \State $\mathcal{P}_{\mathrm{keep}}\gets \mathcal{P}_{\mathrm{keep}}\cup\{p\}$
\EndFor
\State Partition $\mathcal{P}_{\mathrm{keep}}$ into $\{\mathcal{B}_{(L,\tau)}\}$ keyed by $(\text{Lang}(p),\text{page\_type}(p))$
\ForAll{$\mathcal{B}_{(L,\tau)}$}
  \State Sort by $\text{Score}(p)$ descending to get $\mathcal{R}_{(L,\tau)}$
  \State \textbf{Select}: Top-$K$ or threshold $\tau$
\EndFor
\State \Return $\{\mathcal{R}_{(L,\tau)}\}$
\end{algorithmic}
\end{algorithm}

\noindent\textbf{Rule Scoring and Ranking.}
To prioritize representative samples while maintaining language balance, we apply a rule-based scoring mechanism with two filtering steps and a weighted scoring function. 
Each page first passes basic validity and language checks to ensure structural completeness and inclusion of only allowed languages. 
For valid pages, a composite score is then computed as
\[
\text{Score} = a_1S_1 + a_2S_2 + a_3S_3 + a_4S_4 + a_5S_5,
\]
where $S_1$ measures block count, $S_2$ the text-area ratio, $S_3$ the element-type diversity, and $S_4$ and $S_5$ indicate the presence of figures and tables, respectively.
The weights $a_i$ are nonnegative and sum to 100. 
Pages are subsequently grouped by language and page type, ranked by their composite scores, and the top samples in each group are selected for fine-grained annotation, as detailed in Algorithm~\ref{alg1}. 
This stage yields in total $200 \times 11 \times 9 = 19{,}800$ high-quality page-level samples across 11 languages and 9 page types.

\noindent\textbf{Region Correction.}  
The selected pages undergo multimodal region correction to improve structural and semantic accuracy. 
Text blocks are refined using the MLLM to correct OCR errors.
%
Mathematical formulas are re-parsed with UniMERNet \cite{wang2024unimernet}, and tables are adjusted—in a two-step process encompassing structure recognition via the Intsig API\footnote{\url{https://www.textin.com/experience/recognize_table_multipage?from=toolstextin-seo-pc}} and subsequent cell-level correction with MLLM assistance.
This process ensures consistent layout, accurate alignment, and high semantic fidelity across all modalities.
\noindent\textbf{Final Human Verification.} The final stage involves comprehensive human verification to ensure overall annotation quality. 
Annotators first inspect all automatically processed pages and remove those with poor layout integrity, low OCR reliability or content involving political disputes and other sensitive issues.
Remaining pages are then reviewed in detail to validate text–region alignment and correct reading order using the internal verification platform. 
Tables and mathematical formulas are re-rendered and cross-checked through visual inspection with tools such as Tables Generator\footnote{\url{https://www.tablesgenerator.com/}} and latexlive\footnote{\url{https://www.latexlive.com/}} to ensure structural correctness and proper rendering. 
This process guarantees that only high-quality, structurally consistent pages are preserved for the final dataset.

\subsection{TEC-VQA}

\subsubsection{Data Acquisition}

%
Following the pipeline in Fig.~\ref{fig: main} (b), scene images containing rich textual content are gathered from web crawling and manual search across six major domains: \textit{public spaces, consumer places, daily products, work \& study, documents \& IDs, and digital displays}. 
For each domain, we construct extensive keyword lists and translate them into the target languages to retrieve and collect diverse scene text images.
A three-stage filtering process guarantees data quality: (1)~automatic de-duplication and resolution checking using perceptual hashing; (2)~text coverage evaluation by counting OCR-detected characters and text-area ratio; and (3)~language balance enforcement through script classification. 
Only images with legible, naturally embedded text and balanced representation across Southeast Asian scripts are retained. 
This structured filtering yields a visually and linguistically diverse corpus suitable for robust MLLM evaluation. 

\subsubsection{Data Annotation}
\noindent\textbf{Re-rendering for multilingual variants.}
Previous multilingual VQA works expanded datasets by directly translating question–answer pairs across languages. 
Such text-only translation neglects \textit{visual text}, causing \textbf{visual–textual misalignment} when OCR or translation errors alter semantics. 
To address this limitation, the TEC-VQA pipeline preserves visual fidelity by re-rendering textual regions within the images themselves.
For each retained image, layout detection and OCR extraction are first performed to obtain structured text regions and coarse reading order. 
Extracted texts are then \textit{translated and semantically rewritten} into target languages, followed by \textit{font-matched inpainting} to reconstruct the image with consistent typography, font weight, and spatial alignment. 
As shown in Fig.~\ref{fig: main} (b), this approach maintains the visual context while producing realistic multilingual counterparts, ensuring that the visible text language precisely matches subsequent QA generation and eliminating the mismatch common in text-only translations.
\noindent\textbf{MLLM-based QA generation.}
The annotation process adopts a multi-stage generation framework to reduce hallucination and enhance QA diversity. 
For each English-language image, an MLLM (e.g., Qwen3-VL \cite{qwen3technicalreport}) is first prompted to propose multiple question candidates grounded in the recognized text regions. 
Each English question is then translated into Chinese, forming bilingual question pairs that preserve semantics but vary in linguistic form. 
The same MLLM independently answers both the English and Chinese questions by referring to their respective language-specific images. 
A separate judging model compares the generated answers across languages, retaining only those question–answer pairs that show consistent reasoning behavior. 
This multi-round generation and judging mechanism significantly mitigates model hallucination, yielding QA pairs that are semantically precise, cross-lingually aligned, and logically verifiable. 
%

\noindent\textbf{Back-translation and human verification.}
Despite the multi-stage safeguards, potential errors remain due to OCR inaccuracies, re-rendering artifacts, or residual MLLM hallucinations. 
Therefore, all generated QAs are back-translated into the visible text language to validate semantic and syntactic coherence. 
Discrepancies or ambiguities detected in this stage are flagged for human inspection. 
Expert annotators then conduct a comprehensive verification process (see Appendix), which includes: (1)~removing unanswerable or trivial questions; (2)~refining question phrasing and normalizing answer expressions (numbers, units, capitalization); (3)~checking strict alignment between question language and visible text; and (4)~assigning capability labels to each question, categorizing them into five distinct skill dimensions.

%% file: sec/4_dataset_statistics.tex
\section{Dataset Statistics} 
SEA-Vision encompasses \textbf{11 languages}---\textit{English (EN), Chinese (ZH), Vietnamese (VI), Thai (TH), Filipino (FIL), Malay (MS), Indonesian (ID), Lao (LO), Khmer (KM), Burmese (MY), and Portuguese (PT)}---capturing the region’s linguistic diversity. 
The Document Parsing part contains a total of \textbf{15{,}234} pages covering 9 document types (e.g., exam papers, academic articles, forms, receipts) across these 11 languages. Each page is annotated with global attributes such as primary language, layout style (single / multi-column / mixed), and noise indicators (e.g., fuzzy scans, watermarks, or colorful backgrounds). 
At the region level, we provide \textbf{243{,}643} block annotations over 10 semantic categories, including reading order labels. We further tag tables with structural attributes such as merged cells.
Comprehensive statistics, detailing the distributions across page, type, and attribute dimensions, are presented in the Appendix.
The TEC-VQA subset contains \textbf{1{,}839} images and \textbf{7{,}496} question–answer pairs constructed from text-centric scenes and documents spanning the same languages and domains. 
Each QA pair is annotated with one or more capability labels, including text recognition, numerical reasoning, comparison, logical reasoning, and spatial understanding. 
%
%
The full per-language breakdown and capability co-occurrence patterns are deferred to the Appendix.

%% file: sec/5_evaluation.tex
\section{Evaluation Methodology}
\input{Tables/table1_dp}

\input{Tables/table2_dp}



\subsection{Document Parsing Evaluation}

Following OmniDocBench \cite{ouyang2025omnidocbench}, document parsing evaluation is divided into four components: pure text, tables, formulas, and reading order. Each component employs task-appropriate similarity metrics to quantify model performance.

\noindent\textbf{Pure Text.}
For text-only regions, we compute the Normalized Edit Distance~(NED) \cite{lcvenshtcin1966binary} to assess textual reconstruction quality. The final score is obtained by averaging NED across all samples.

\noindent\textbf{Tables.}
Tables are converted to HTML, then evaluated using TEDS \cite{zhong2020teds} and NED.

\noindent\textbf{Formulas.}
Evaluation is based on Character Detection Matching metric~(CDM) \cite{wang2024cdm}, NED, and BLEU \cite{papineni2002bleu}.

\noindent\textbf{Reading Order.}
Reading order evaluation relies on the NED metric.

\subsection{TEC-VQA Evaluation}

Evaluation for TEC-VQA is primarily conducted using accuracy. This metric is defined by the proportion of questions where the predicted answer is deemed correct if it precisely encompasses at least one of the reference answers. This containment criterion effectively captures the fidelity of textual content recognition within visual contexts, inherently tolerating minor output variations.

%% file: Tables/table1_dp.tex
\begin{table*}[t]
\centering
\large
\setlength{\tabcolsep}{6pt}      
\renewcommand{\arraystretch}{1.25} 

\resizebox{\textwidth}{!}{%
\begin{tabular}{l|l|ccccccccccc|c}
\hline
\textbf{Method Type}    & \textbf{Methods}        & \textbf{EN}    & \textbf{FIL}   & \textbf{ID}    & \textbf{KM}    & \textbf{LO}    & \textbf{MS}    & \textbf{MY}    & \textbf{PT}    & \textbf{TH}    & \textbf{VI}    & \textbf{ZH}    & \textbf{Avg.} \\
\hline
\multirow{5}{*}{\textbf{Pipeline Models}} & \textbf{MinerU2.5} \cite{niu2025mineru2}      & 0.187 & 0.106 & 0.194 & 0.693 & 0.730 & 0.213 & 0.520 & 0.209 & 0.592 & 0.278 & 0.218 & 0.358\\
      & \textbf{Dolphin-1.5} \cite{feng2025dolphin}    & 0.156 & 0.089 & 0.166 & 0.689 & 0.736 & 0.165 & 0.479 & 0.150 & 0.549 & 0.167 & 0.204 & 0.323\\
      & \textbf{MonkeyOCR-pro-1.2B} \cite{li2025monkeyocr}      & 0.128 & 0.094 & 0.212 & 0.668 & 0.697 & 0.209 & 0.468 & 0.207 & 0.574 & 0.457 & 0.162 & 0.352\\
      & \textbf{MonkeyOCR-pro-3B} \cite{li2025monkeyocr}        & 0.135 & 0.092 & 0.171 & 0.659 & 0.688 & 0.159 & 0.488 & 0.157 & 0.578 & 0.370 & 0.151 & 0.331\\
      & \textbf{PaddleOCR-VL} \cite{cui2025paddleocr}  & \textbf{0.108} & 0.065 & 0.099 & 0.634 & 0.648 & 0.087 & 0.456 & 0.107 & 0.133 & 0.148 & \textbf{0.131} & 0.238\\
\hline
\multirow{3}{*}{\textbf{Expert Models}}   & \textbf{POINTS-Reader} \cite{liu2025points} & 0.121 & \textbf{0.046} & \textbf{0.092} & 0.845 & 0.815 & \textbf{0.075} & 0.755 & 0.115 & 0.709 & 0.184 & 0.170 & 0.357\\
      & \textbf{DeepSeek-OCR} \cite{wei2025deepseekocr}  & 0.197 & 0.073 & 0.113 & 0.673 & 0.252 & 0.133 & 0.499 & 0.151 & 0.329 & 0.216 & 0.172 & 0.255\\
      & \textbf{dots.ocr} \cite{dots.ocr}      & 0.144 & 0.065 & 0.116 & 0.311 & 0.386 & 0.096 & 0.313 & 0.121 & 0.158 & 0.152 & 0.181 & 0.186\\
\hline
\multirow{5}{*}{\textbf{General Models}}  & \textbf{InternVL3.5-38B} \cite{wang2025internvl3_5} & 0.416 & 0.450 & 0.483 & 0.734 & 0.752 & 0.503 & 0.618 & 0.521 & 0.695 & 0.587 & 0.675 & 0.585\\
      & \textbf{Qwen2.5-VL-72B} \cite{Qwen2.5-VL} & 0.159 & 0.080 & 0.115 & 0.783 & 0.612 & 0.091 & 0.579 & 0.149 & 0.112 & 0.122 & 0.188 & 0.272\\
      & \textbf{Qwen3-VL-32B} \cite{qwen3technicalreport}  & 0.133 & 0.067 & 0.098 & 0.727 & 0.406 & 0.085 & 0.479 & \textbf{0.099} & 0.114 & \textbf{0.107} & 0.160 & 0.225\\
      & \textbf{GPT4o} \cite{hurst2024gpt} & 0.197 & 0.097 & 0.189 & 0.611 & 0.610 & 0.162 & 0.423 & 0.235 & 0.218 & 0.244 & 0.461 & 0.313\\
      & \textbf{Gemini2.5-Pro} \cite{comanici2025gemini} & 0.154 & 0.083 & 0.124 & \textbf{0.278} & \textbf{0.195} & 0.122 & \textbf{0.214} & 0.134 & \textbf{0.107} & 0.132 & 0.203 & \textbf{0.159}\\
\hline
\multicolumn{2}{c|}{\textbf{Avg.}}        & 0.172 & 0.108 & 0.167 & 0.639 & 0.579 & 0.162 & 0.484 & 0.181 & 0.374 & 0.243 & 0.236 & 0.304  \\
\hline
\end{tabular}
}
\caption{\normalfont End-to-end document parsing performance on SEA-Vision, reported as Normalized Edit Distance (NED↓) across 11 languages. Best scores per language are \textbf{bolded}.}
\label{table1: dp_lang}
\end{table*}

%% file: Tables/table2_dp.tex
\begin{table*}[t]
\centering
\large
\setlength{\tabcolsep}{6pt}      
\renewcommand{\arraystretch}{1.25} 

\resizebox{0.95\textwidth}{!}{%
\begin{tabular}{l|l|ccccccccc|c}
\hline
\multicolumn{1}{c|}{\textbf{Method Type}}  & \multicolumn{1}{c|}{\textbf{Methods}} & \textbf{\makecell{Academic\\Paper}} & \textbf{Book} & \textbf{\makecell{Exam\\Paper}} & \textbf{Magazine} & \textbf{Newspaper} & \textbf{Note} & \textbf{\makecell{Research\\Report}} & \textbf{Slide} & \textbf{Textbook} & \textbf{Avg.} \\
\hline
\multirow{5}{*}{\textbf{Pipeline Models}} & \textbf{MinerU2.5} \cite{niu2025mineru2} & 0.301 & 0.348& 0.369      & 0.301    & 0.446     & 0.289& 0.335  & 0.303 & 0.360    & 0.339\\
      & \textbf{Dolphin-1.5} \cite{feng2025dolphin}       & 0.235 & 0.304& 0.377      & 0.221    & 0.158     & 0.319& 0.281  & 0.297 & 0.285    & 0.275\\
      & \textbf{MonkeyOCR-pro-1.2B} \cite{li2025monkeyocr} & 0.272 & 0.348& 0.391      & 0.340    & 0.291     & 0.353& 0.340  & 0.328 & 0.349    & 0.335\\
      & \textbf{MonkeyOCR-pro-3B} \cite{li2025monkeyocr}  & 0.245 & 0.325& 0.371      & 0.478    & 0.232     & 0.361& 0.314  & 0.315 & 0.326    & 0.330\\
      & \textbf{PaddleOCR-VL}  \cite{cui2025paddleocr}     & 0.156 & 0.205& 0.265      & 0.148    & 0.119     & 0.255& 0.214  & 0.204 & 0.217    & 0.198\\
\hline
\multirow{3}{*}{\textbf{Expert Models}}   & \textbf{POINTS-Reader}  \cite{liu2025points}    & 0.240 & 0.329& 0.328      & 0.208    & 0.318     & 0.374& 0.318  & 0.289 & 0.317    & 0.302\\
      & \textbf{DeepSeek-OCR} \cite{wei2025deepseekocr}      & 0.194 & 0.269& 0.253      & 0.200    & 0.313     & 0.324& 0.219  & 0.221 & 0.268    & 0.251\\
      & \textbf{dots.ocr} \cite{dots.ocr} & \textbf{0.141} & \textbf{0.160} & \textbf{0.206}      & 0.204    & 0.192     & 0.250& 0.169  & 0.195 & 0.198    & 0.190\\
\hline
\multirow{5}{*}{\textbf{General Models}}  & \textbf{InternVL3.5-38B} \cite{wang2025internvl3_5}     & 0.604 & 0.586& 0.549      & 0.641    & 0.720     & 0.519& 0.598  & 0.427 & 0.525    & 0.574\\
      & \textbf{Qwen2.5-VL-72B} \cite{Qwen2.5-VL}   & 0.195 & 0.229& 0.254      & 0.180    & 0.211     & 0.308& 0.242  & 0.252 & 0.255    & 0.236\\
      & \textbf{Qwen3-VL-32B} \cite{qwen3technicalreport}      & 0.163 & 0.175& 0.243      & 0.144    & 0.212     & 0.330& 0.202  & 0.225 & 0.220    & 0.213\\
      & \textbf{GPT4o} \cite{hurst2024gpt}    & 0.283 & 0.333& 0.249      & 0.273    & 0.436     & 0.269& 0.299  & \textbf{0.184} & 0.276    & 0.289\\
      & \textbf{Gemini2.5-Pro} \cite{comanici2025gemini}     & 0.143 & 0.162& 0.217      & \textbf{0.123}    & \textbf{0.116}     & \textbf{0.099}& \textbf{0.145}  & 0.278 & \textbf{0.146}    & \textbf{0.159}\\
\hline
\multicolumn{2}{c|}{\textbf{Avg.}}   & 0.244 & 0.290& 0.313      & 0.266    & 0.289     & 0.312& 0.283  & 0.271 & 0.288    & 0.284\\
\hline
\end{tabular}
}
\caption{\normalfont End-to-end document parsing performance on SEA-Vision, reported as Normalized Edit Distance (NED↓) across 9 page types. Best scores per page type are \textbf{bolded}.
}
\label{table2: dp_page}
\end{table*}

%% file: sec/6_benchmark.tex
\section{Experiments}

\input{Tables/table3_VQA}

\subsection{Baselines}
\noindent\textbf{Document Parsing.}
We group baselines into three categories based on their modeling paradigm.
(1) \emph{Pipeline Models} couple a layout detector with task-specific recognizers for OCR, table recognition, and formula recognition, and then fuse all outputs into a structured representation. 
(2) \emph{Expert Models} are specialized vision–language models for document parsing.
(3) \emph{General Models} are open- or closed-source MLLMs that are not specifically trained for document parsing but can perform it in a zero-shot manner.
All baselines are evaluated in an end-to-end setting using their official checkpoints and configurations.
\noindent\textbf{TEC-VQA.} For this task, we consider two groups of MLLMs: open-source general MLLMs and closed-source general MLLMs. 
To ensure a fair comparison, all models are assessed in a unified, zero-shot setting using the same prompt template and the answer accuracy is computed directly from their raw textual outputs without any task-specific fine-tuning.
\subsection{Overall Results}
For document parsing, we report NED$\downarrow$ where lower is better. For TEC-VQA (Text-centric Visual Question Answering), we report answer accuracy$\uparrow$ where higher is better.

\subsubsection{Document Parsing}
\noindent\textbf{Performance Gaps Across Languages.}
As shown in Tab. \ref{table1: dp_lang} across all 13 baselines and 11 languages, the General Model category—led by Gemini 2.5 Pro—achieves state-of-the-art performance with the lowest average NED of 0.159, outperforming the best Expert Model, dots.ocr (0.186), and the strongest Pipeline Model, PaddleOCR-VL (0.238). Gemini 2.5 Pro further ranks first in four languages, excelling particularly on low-resource Southeast Asian scripts such as Khmer (0.278), Lao (0.195), Burmese(0.214) and Thai (0.107), underscoring its strong multilingual generalization in complex document parsing.

Conversely, PaddleOCR-VL achieves the best results on the two high-resource languages—English (0.108) and Chinese (0.131)—showing that specialized pipelines can still outperform MLLMs when ample language-specific data and tailored modules are available. Overall, these results highlight MLLMs’ superior cross-lingual robustness while indicating that expert or pipeline systems may still hold advantages in well-resourced settings.

%
%
%

\noindent\textbf{Performance Gaps Across Page Types.}
Referring to Tab. \ref{table2: dp_page}, Expert Models excel on fixed-layout documents with dots.ocr achieving top scores on Academic Paper (0.141) and Book (0.160). In contrast, General Models show markedly higher robustness on complex and irregular layouts. Gemini 2.5 Pro sets new benchmarks in four categories, including the challenging Magazine (0.123) and Newspaper (0.116) types, demonstrating superior layout generalization and real-world applicability. 

Our benchmark highlights a functional dichotomy: Expert Models remain strong for structured, layout-consistent extraction, while General Models provide superior accuracy and resilience in visually complex settings such as Newspapers and Magazines. Future work should focus on enhancing complex document structure understanding in Expert Models or improving fine-grained localization in General Models to further reduce the performance gap across diverse document types.
%
%
%

\noindent\textbf{Other sub-tasks.}
Beyond page-level scores, we further break down document parsing into pure text, tables, formulas, and reading-order evaluation. Due to space limitations, the detailed analysis is provided in the Appendix.
%

\subsubsection{TEC-VQA}
\noindent\textbf{Overall performance and multilingual trends.}
On TEC-VQA, all models perform substantially worse than on document parsing. 
Averaged over the 11 languages, the overall accuracy is only 32.36\%, showing that text-centric visual question answering in SEA-Vision remains far from solved. 
Closed-source systems~(GPT-4o and Gemini~2.5~Pro) reach an average of 37.49\%, while the open-source group averages 31.34\%. 
The best-performing model, Qwen3-VL-32B, attains 40.14\% accuracy, followed closely by Gemini~2.5~Pro with 39.49\%, whereas smaller models such as DeepSeek-VL2-3B remain around 23.01\%. 
The language-level averages also show a clear trend: English~(EN) reaches 62.85\% accuracy and Chinese~(ZH) 49.33\%, but accuracy drops to 43.61\% on Indonesian~(ID), and further down to 13.34\%, 11.26\%, and 8.17\% on Lao~(LO), Khmer~(KM), and Burmese~(MY), respectively. 
This large spread confirms that current MLLMs still struggle with robust text-centric reasoning on many Southeast Asian languages.



%% file: Tables/table3_VQA.tex
\begin{table*}[t]
\centering
\large
\setlength{\tabcolsep}{6pt}      
\renewcommand{\arraystretch}{1.25} 
\resizebox{0.9\textwidth}{!}{%
\begin{tabular}{l|ccccccccccc|c}
\hline
  & \textbf{EN}    & \textbf{ZH}    & \textbf{ID}    & \textbf{PT}    & \textbf{MS}    & \textbf{VI}    & \textbf{TH}    & \textbf{FIL}   & \textbf{LO}    & \textbf{KM}    & \textbf{MY}    & \textbf{Avg.} \\
\hline
\multicolumn{13}{l}{\textit{\textbf{Closed-Source}}}   \\
\hline
\textbf{GPT4o} \cite{hurst2024gpt}   & 60.97& 42.80& 49.02& 31.43& 30.09& 25.76& 28.57& 21.84& 19.68& 16.92& 14.25& 35.49    \\
\textbf{Gemini2.5-Pro} \cite{comanici2025gemini} & 69.14& 56.40& 47.98& 33.95& 29.44& 26.46& \textbf{29.89} & 20.10& \textbf{27.49} & \textbf{28.92} & \textbf{25.70} & 39.49    \\
\hline
\multicolumn{13}{l}{\textit{\textbf{Open-source}}}\\
\hline
\textbf{Qwen2.5-VL-72B} \cite{Qwen2.5-VL}   & 71.09& 57.80& 49.28& 35.29& 30.41& 26.23& 24.06& 20.35& 18.06& 12.00& 8.66 & 37.43    \\
\textbf{Qwen3-VL-32B} \cite{qwen3technicalreport} & \textbf{71.69} & \textbf{64.40} & \textbf{49.67} & \textbf{37.82} & \textbf{32.58} & \textbf{28.81} & 28.48& \textbf{24.07} & 22.64& 15.69& 11.73& \textbf{40.14}    \\
\textbf{InternVL3-78B} \cite{zhu2025internvl3} & 58.35& 47.40& 42.50& 30.42& 27.16& 22.13& 17.48& 18.11& 11.59& 10.46& 5.59 & 30.94    \\
\textbf{InternVL3.5-38B} \cite{wang2025internvl3_5} & 60.67& 51.40& 46.28& 32.44& 29.87& 26.93& 16.73& 21.84& 12.67& 10.77& 9.22 & 33.38    \\
\textbf{MiniCPM-V-4.5} \cite{yu2025minicpm4_5} & 63.30& 51.40& 44.72& 30.76& 25.97& 21.08& 12.78& 18.86& 11.59& 9.23 & 5.87 & 31.40    \\
\textbf{MiniCPM-V-2.6} \cite{yao2024minicpm2_6} & 58.28& 41.80& 35.72& 26.55& 23.38& 15.22& 8.93 & 13.90& 5.39 & 6.15 & 1.40 & 26.16    \\
\textbf{Ovis2-34B}  \cite{brandt2008ovis}  & 67.27& 52.40& 47.20& 33.78& 31.28& 25.29& 17.11& 19.35& 12.40& 11.38& 6.98 & 34.63    \\
\textbf{DeepSeek-VL2-27B} \cite{wu2024deepseek} & 54.76& 47.40& 39.77& 24.87& 22.51& 15.81& 11.75& 14.89& 3.23 & 4.00 & 1.68 & 26.41    \\
\textbf{DeepSeek-VL2-3B} \cite{wu2024deepseek}  & 54.31& 39.40& 30.77& 23.87& 18.29& 12.18& 8.18 & 10.92& 2.70 & 1.54 & 1.68 & 23.01    \\
\textbf{LLaVA-OV-1.5-8B} \cite{an2025llava} & 64.42& 39.40& 40.42& 28.07& 26.95& 20.37& 11.09& 18.11& 12.67& 8.00 & 5.31 & 29.88    \\
\hline

\textbf{Avg.}    & 62.85& 49.33& 43.61& 30.77& 27.33& 22.19& 17.92& 18.53& 13.34& 11.26& 8.17 & 32.36 \\  
\hline
\end{tabular}
}
\caption{\normalfont Performance of the leading closed- and open-source MLLMs on the TEC-VQA. The best results of each language are \textbf{bolded}.}
\label{table1:sota}
\end{table*}

%% file: sec/7_conclusion.tex
\section{Conclusion}
This paper addresses the lack of realistic multilingual benchmarks for document and scene text understanding in Southeast Asia. We introduce SEA-Vision, a unified dataset covering 11 languages that supports both Document Parsing and TEC-VQA, constructed through a hybrid pipeline that combines automatic annotation with native-speaker verification. Together with a unified evaluation protocol, SEA-Vision enables systematic and fair comparison of leading models in challenging multilingual settings. Task-level and language-level analyses reveal clear gaps between high-resource and low-resource languages. These findings guide targeted model optimization and lay a practical foundation for more robust and effective multilingual  document and scene text understanding systems.

%% file: Appendix/Appendix.tex
\section{Dataset Statistics}
As outlined in Section 4 of the main text, this appendix expands the statistical analysis of SEA-Vision. It provides additional details for the Document Parsing subset, the distribution of capability labels in the TEC-VQA portion, and a co-occurrence analysis of reasoning skills.

\subsection{Document Parsing Statistics}

\noindent\textbf{Language-wise distribution.}
The Document Parsing subset spans 11 languages, with page counts ranging from 636 to 1,678 (Table \ref{tab:dp_lang_stats}). The comparatively small Chinese portion (636 pages) results from strict safety filtering: many candidate documents contained politically sensitive content and were removed. Lao and Khmer also remain smaller due to the scarcity of high-quality public materials in these languages.
In contrast, English, Vietnamese, Malay, Indonesian, Myanmar, and Portuguese each contribute 1.5k–1.7k pages, reflecting richer public sources.
Layout complexity varies substantially across languages: English and Chinese pages contain dense structures with over 25 blocks on average, whereas Lao, Khmer, Thai, and Myanmar typically exhibit simpler layouts of 8–10 blocks. Content density is also uneven: Chinese pages show the highest formula proportion (16.5\%), consistent with their larger share of technical and educational materials.

\begin{table}[t]
\centering
\Large
\setlength{\tabcolsep}{6pt}      
\caption{Per-language statistics for the Document Parsing subset of SEA-Vision. We report the number of pages, average number of blocks per page, average text-area ratio, and the proportions of pages with tables and formulas.}
\label{tab:dp_lang_stats}
\resizebox{0.48\textwidth}{!}{
\begin{tabular}{lrrrrr}
\hline
Language & \#Pages & Avg. blocks/page & Avg. text-area ratio & Pages with tables (\%) & Pages with formulas (\%) \\
\hline
EN  & 1585 & 25.71 & 63.08 & 15.65 &  6.56 \\
ZH  &  636 & 28.11 & 62.59 & 23.27 & 16.50 \\
VI  & 1678 & 19.69 & 60.13 & 11.50 &  5.78 \\
TH  & 1506 & 10.53 & 54.38 &  6.57 &  1.26 \\
FIL & 1265 & 12.35 & 49.40 &  8.06 &  0.70 \\
MS  & 1580 & 14.12 & 49.50 &  9.68 &  0.89 \\
ID  & 1532 & 17.59 & 53.79 & 18.73 &  6.33 \\
LO  &  914 &  8.86 & 49.85 &  3.71 &  0.11 \\
KM  & 1331 &  8.49 & 52.50 &  3.38 &  0.38 \\
MY  & 1575 &  9.68 & 52.56 &  5.14 &  0.76 \\
PT  & 1632 & 22.79 & 60.90 & 12.81 &  2.80 \\
\hline
\end{tabular}
}
\end{table}

\noindent\textbf{Page-type-wise distribution.}
We categorize pages into nine representative document types covering diverse real-world formats, including exam papers, academic articles, books, magazines, newspapers, handwritten notes, research reports, slides, and textbooks (Table \ref{tab:dp_type_stats}).
Academic, textbook, and book pages constitute the core of the dataset and remain relatively well balanced across most languages. By contrast, Chinese exam papers, magazines, and newspapers appear less frequently due to the removal of safety-sensitive materials. Magazine and newspaper pages are heavily concentrated in high-resource languages such as English, Vietnamese, Malay, Indonesian, and Portuguese, with minimal representation for Lao and Khmer.
Handwritten notes and slides are included for all languages but remain notably sparse in low-resource ones, making cross-lingual evaluation on informal or noisy layouts more challenging.

\begin{table}[t]
\centering
\caption{Page-type statistics for the Document Parsing subset. Each entry denotes the number of pages for a given page type and language.}
\label{tab:dp_type_stats}
\scriptsize
\resizebox{0.45\textwidth}{!}{
\begin{tabular}{lrrrrrrrrrrr}
\hline
Page Type& EN  & ZH  & VI  & TH  & FIL & MS  & ID  & LO  & KM  & MY  & PT  \\
\hline
Academic Literature & 164 & 178 & 185 & 179 & 161 & 168 & 140 & 61  & 192 & 199 & 176 \\
Book  & 187 & 64  & 186 & 193 & 187 & 187 & 187 & 200 & 199 & 171 & 193 \\
Textbook   & 183 & 47  & 184 & 179 & 177 & 187 & 177 & 135 & 188 & 192 & 182 \\
Exam Paper  & 133 & 9   & 184 & 157 & 141 & 147 & 146 & 129 & 148 & 166 & 180 \\
Magazine & 189 & 56  & 200 & 181 & 65  & 196 & 194 & 4   & 20  & 92  & 196 \\
Newspaper& 175 & 124 & 185 & 49  & 55  & 187 & 186 & 3   & 6   & 188 & 171 \\
Note  & 193 & 39  & 190 & 196 & 190 & 190 & 198 & 125 & 198 & 198 & 191 \\
Slide  & 188 & 22  & 196 & 196 & 132 & 160 & 169 & 70  & 196 & 195 & 184 \\
Research Report     & 173 & 97  & 168 & 176 & 157 & 158 & 135 & 187 & 184 & 174 & 159 \\
\hline
\end{tabular}
}
\end{table}

\subsection{TEC-VQA Statistics}

\noindent\textbf{Per-language question–answer distribution.}
The Text-Centric Visual Question Answering (TEC-VQA) subset contains 1,839 images and 7,496 question–answer (QA) pairs, spanning the same 11 languages as the Document Parsing subset. As shown in Table~\ref{tab:vqa_lang_stats}, we report, for each language, the number of images, the number of QA pairs, and the average lengths of questions and answers in tokens. 
It can be seen that, for each low-resource language, TEC-VQA still maintain non-trivial coverage to support multi-lingual evaluation.

\begin{table}[t]
\centering
\caption{Per-language statistics for the TEC-VQA subset, including the number of images, the number of QA pairs, and the average question and answer lengths.}
\label{tab:vqa_lang_stats}
\resizebox{0.45\textwidth}{!}{
\begin{tabular}{lrrrr}
\hline
Language & \#Images & \#QA pairs & Avg Q length & Avg A length \\
\hline
EN  & 267& 1335& 72.74& 31.00\\
ZH  & 100& 500& 21.72& 12.81\\
VI  & 225& 854& 68.35& 47.19\\
TH  & 272& 1064& 57.01& 35.31\\
FIL & 115& 403& 82.61& 86.71\\
MS  & 242& 924& 73.54& 38.29\\
ID  & 200& 767& 71.07& 37.95\\
LO  & 89& 371& 58.33& 43.65\\
KM  & 85& 325& 66.37& 28.80\\
MY  & 242& 358& 68.08& 32.96\\
PT  & 152& 595& 73.37& 46.09\\
\hline
\end{tabular}
}
\end{table}

\noindent\textbf{Capability distribution.}
Each TEC-VQA question is labeled with one or more of five reasoning skill categories: text recognition (TR), numerical calculation (NC), comparative analysis (CA), logical reasoning (LR), and spatial understanding (SU). These labels indicate the primary capabilities required to answer the question. Text recognition questions involve directly reading a text snippet from the image. Numerical calculation questions require arithmetic or counting based on textual content. Comparative analysis questions compare two or more values (for example, selecting the larger number or the earlier date). Logical reasoning questions refers to those requiring logical judgment, such as truth and falsehood, and AND/OR/NOT relationships. Spatial understanding questions depend on layout or positional relations of text elements (for example, “Which item is at the top of the page?”). The overall distribution of these capability labels is summarized in Table~\ref{tab:vqa_cap_stats}.

\begin{table}[t]
\centering
\caption{Distribution of TEC-VQA capability categories in SEA-Vision, reported as the number and proportion of QA pairs attributed to each capability.}
\label{tab:vqa_cap_stats}
\resizebox{0.45\textwidth}{!}{
\begin{tabular}{lrr}
\hline
Capability category & \#QA pairs & Proportion (\%) \\
\hline
Text recognition& 5573& 74.35\\
Numerical calculation   & 5065& 67.57\\
Comparative analysis    & 4306& 57.44\\
Logical reasoning       & 459& 6.12\\
Spatial understanding   & 209& 2.79\\
\hline
\end{tabular}
}
\end{table}

\begin{figure}[t]
\centering
\includegraphics[width=0.48\textwidth]{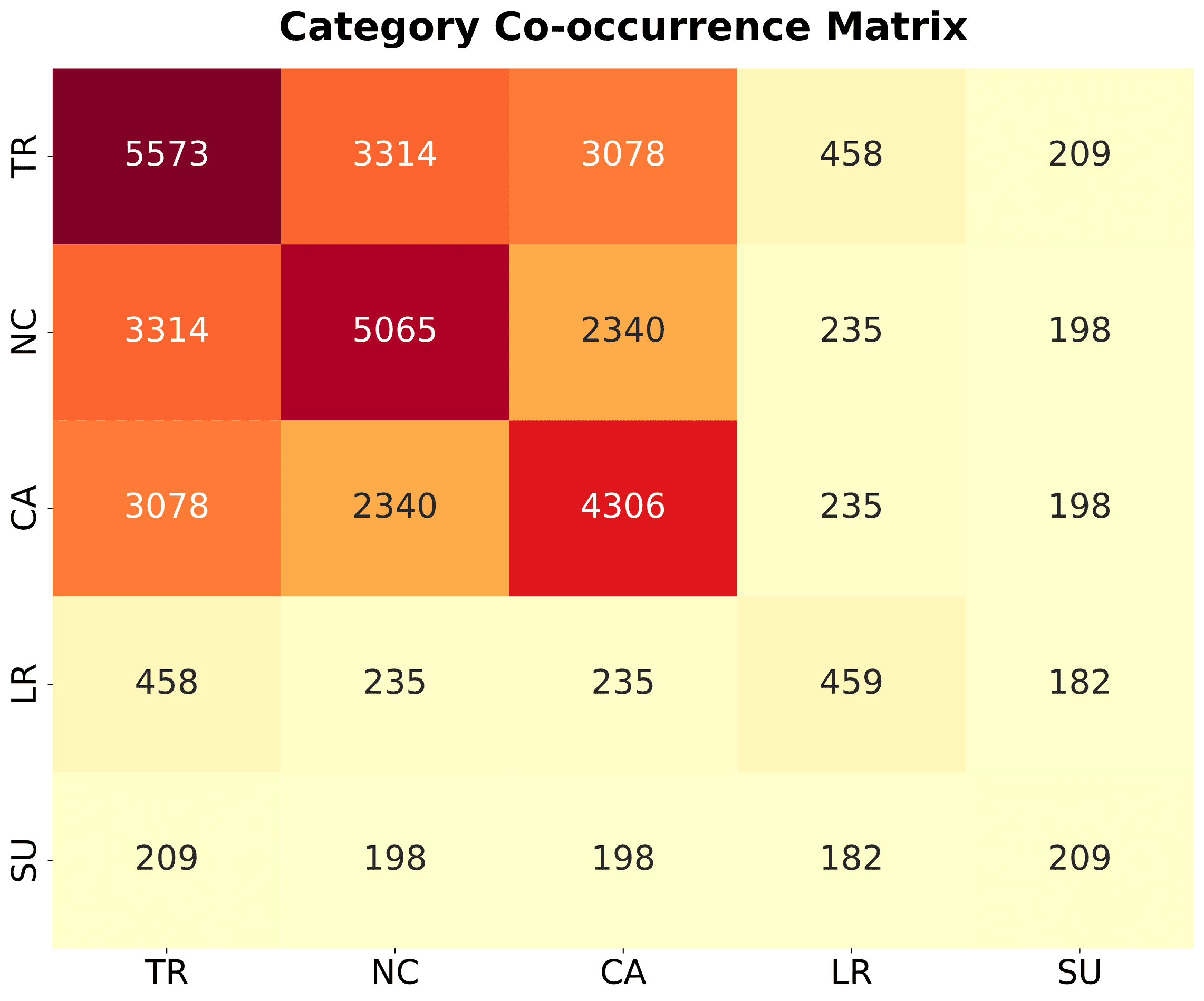}
\caption{Heatmap of the TEC-VQA capability category co-occurrence matrix. Color intensity indicates the frequency of QA pairs annotated with each capability combination. }
\label{fig:vqa_cap_cooc_heatmap}
\end{figure}

\noindent\textbf{Capability co-occurrence analysis.}
Many TEC-VQA questions require multiple capabilities simultaneously (for example, reading several numbers and comparing them after a simple calculation). To analyze such combinations in TEC-VQA, we compute the co-occurrence matrix over the capability labels. Each off-diagonal entry counts QA pairs in Fig.~\ref{fig:vqa_cap_cooc_heatmap} annotated with the corresponding capability combination (for example, ``2430" represents the number of questions requiring text recognition and comparative analysis simultaneously). We observe that numerical calculation frequently co-occurs with text recognition and comparative analysis, reflecting that images contain a wealth of statistical information, such as tabulated bills, receipts, or timetables, which often require reading, comparing values, and further calculation. Logical reasoning and spatial understanding appear less frequently overall but tend to co-occur with text recognition in more complex layout-dependent questions.


\section{Additional Quantitative Results}
This appendix provides additional quantitative analyses that decompose the overall results in Section~5 into finer-grained sub-tasks and data subsets. We first present Document Parsing performance on pure text, tables, formulas, and reading order, and then provide a detailed breakdown of TEC-VQA by domain and language/resource level. We use the same metrics as in the main paper: Normalized Edit Distance (NED, Normalized Edit Distance), Tree Edit Distance-based Similarity (TEDS, Tree Edit Distance-based Similarity), Character Detection Matching metric (CDM, Character Detection Matching metric), and Bilingual Evaluation Understudy (BLEU, Bilingual Evaluation Understudy).

\subsection{Document Parsing Sub-Task Results}

\subsubsection{Pure Text}
Table \ref{tab:pure_text_lang} shows clear cross-language trends across model families. Latin- and Chinese-based languages exhibit relatively low NED—strong models typically stay around 0.05–0.12 on EN, FIL, ID, and ZH—whereas Khmer, Lao, Thai, and Burmese remain considerably more challenging, with many baselines exceeding 0.80 and some pipeline systems reaching 0.96–0.98. Pipeline and expert models perform well on familiar scripts (e.g., POINTS-Reader at 0.05 on EN and 0.04 on ID) but degrade sharply on low-resource ones (e.g., 0.975 on LO and 0.981 on KM). General models show more stable cross-lingual performance: Qwen3-VL-32B maintains low NED across multiple high-resource languages, and Gemini 2.5-Pro achieves the best overall average (0.129) while reducing errors on difficult scripts such as Khmer (0.312), Lao (0.150), and Burmese (0.214). These results reflect pronounced and persistent performance gaps across languages in pure-text recognition.

\begin{table*}[t]
\centering
\setlength{\tabcolsep}{6pt}
\renewcommand{\arraystretch}{1.25}

\resizebox{\textwidth}{!}{%
\begin{tabular}{l|l|ccccccccccc|c}
\hline
\textbf{Method Type}& \textbf{Methods}   & \multicolumn{1}{c}{\textbf{EN}} & \multicolumn{1}{c}{\textbf{FIL}} & \multicolumn{1}{c}{\textbf{ID}} & \multicolumn{1}{c}{\textbf{KM}} & \multicolumn{1}{c}{\textbf{LO}} & \multicolumn{1}{c}{\textbf{MS}} & \multicolumn{1}{c}{\textbf{MY}} & \multicolumn{1}{c}{\textbf{PT}} & \multicolumn{1}{c}{\textbf{TH}} & \multicolumn{1}{c}{\textbf{VI}} & \multicolumn{1}{c|}{\textbf{ZH}} & \multicolumn{1}{c}{\textbf{Avg.}} \\
\hline
\multirow{5}{*}{\textbf{Pipeline Models}} & \textbf{MinerU2.5} & 0.234  & 0.186& 0.248  & 0.969  & 0.975  & 0.223  & 0.858  & 0.255  & 0.972  & 0.469  & 0.307  & 0.503 \\
& \textbf{Dolphin-1.5}  & 0.061  & 0.065& 0.063  & 0.962  & 0.932  & 0.068  & 0.688  & 0.039  & 0.701  & 0.108  & 0.097  & 0.321 \\
& \textbf{MonkeyOCR-pro-1.2B}      & 0.059  & 0.07 & 0.241  & 0.985  & 0.985  & 0.248  & 0.806  & 0.24& 0.987  & 0.878  & 0.091  & 0.51  \\
& \textbf{MonkeyOCR-pro-3B}& 0.054  & 0.057& 0.121  & 0.971  & 0.974  & 0.097  & 0.756  & 0.059  & 0.972  & 0.721  & 0.087  & 0.435 \\
& \textbf{PaddleOCR-VL} & 0.044  & 0.046& 0.036  & 0.966  & 0.964  & 0.041  & 0.787  & 0.039  & 0.081  & 0.172  & 0.083  & 0.268 \\
\hline
\multirow{3}{*}{\textbf{Expert Models}}   & \textbf{POINTS-Reader}& 0.05& 0.022& 0.04& 0.975  & 0.981  & 0.032  & 0.909  & 0.051  & 0.857  & 0.202  & 0.117  & 0.363 \\
& \textbf{DeepSeek-OCR} & 0.118  & 0.037& 0.056  & 0.656  & 0.163  & 0.053  & 0.499  & 0.082  & 0.244  & 0.171  & 0.147  & 0.2   \\
& \textbf{dots.ocr}  & 0.05& 0.033& 0.05& 0.333  & 0.335  & 0.04& 0.244  & 0.039  & 0.084  & 0.1 & 0.103  & 0.118 \\
\hline
\multirow{5}{*}{\textbf{General Models}}  & \textbf{InternVL3.5-38B}& 0.264  & 0.413& 0.414  & 0.962  & 0.97& 0.419  & 0.787  & 0.453  & 0.943  & 0.577  & 0.761  & 0.605 \\
& \textbf{Qwen2.5-VL-72B-Instruct} & 0.064  & 0.057& 0.057  & 0.922  & 0.897  & 0.054  & 0.761  & 0.043  & 0.094  & 0.069  & 0.118  & 0.256 \\
& \textbf{Qwen3-VL-32B-Instruct}   & 0.055  & 0.053& 0.056  & 0.925  & 0.505  & 0.057  & 0.65& 0.04& 0.113  & 0.076  & 0.086  & 0.223 \\
& \textbf{GPT4o}     & 0.124  & 0.091& 0.115  & 0.798  & 0.797  & 0.11& 0.577  & 0.147  & 0.323  & 0.214  & 0.595  & 0.303 \\
& \textbf{Gemini2.5-Pro}& 0.068  & 0.086& 0.079  & 0.312  & 0.15& 0.072  & 0.336  & 0.051  & 0.102  & 0.098  & 0.13& 0.129 \\
\hline
\multicolumn{2}{c|}{\textbf{Avg.}}  & 0.096  & 0.094& 0.121  & 0.826  & 0.741  & 0.116  & 0.666  & 0.118  & 0.498  & 0.297  & 0.209  & 0.326 \\    \hline 
\end{tabular}

}
\caption{\normalfont Pure-text region performance on SEA-Vision, reported as Normalized Edit Distance (NED↓) across 11 languages.}
\label{tab:pure_text_lang}
\end{table*}

\subsubsection{Tables}
Table \ref{tab:tables_big} shows that table parsing performance varies substantially across languages and model families. For high-resource Latin and Chinese scripts, strong models achieve low text errors (e.g., NED around 0.05–0.12 for EN, ID, MS, ZH) and high structural fidelity (TEDS above 0.90 for several baselines). In contrast, table regions in Khmer and Lao remain challenging for nearly all systems, with NED often exceeding 0.60 and TEDS dropping below 0.35, reflecting frequent structural mismatches and cell-level OCR mistakes. Pipeline and expert models maintain strong structure-text consistency on familiar scripts—for instance, PaddleOCR-VL reaches TEDS 0.91–0.93 on EN/ID and NED around 0.06—but degrade sharply on low-resource scripts (e.g., TEDS 0.23–0.32 on KM/LO). General models exhibit more stable behavior across languages, with Qwen3-VL-32B and Gemini 2.5-Pro delivering the most balanced results: Gemini achieves TEDS above 0.92 on several high-resource languages and lowers NED to 0.40–0.41 on Khmer and Lao, outperforming other general models. Overall, the table results show a pronounced gap between high-resource and low-resource scripts, driven jointly by structural ambiguity and noise in complex table regions.



\begin{table*}[t]
\centering
\large
\setlength{\tabcolsep}{4pt}
\renewcommand{\arraystretch}{1.25}

\resizebox{\textwidth}{!}{
\begin{tabular}{l|l|
cc|cc|cc|cc|cc|cc|cc|cc|cc|cc|cc|cc}
\hline
\multirow{2}{*}{\textbf{Method Type}} &
\multirow{2}{*}{\textbf{Methods}} &
\multicolumn{2}{c|}{\textbf{EN}} &
\multicolumn{2}{c|}{\textbf{FIL}} &
\multicolumn{2}{c|}{\textbf{ID}} &
\multicolumn{2}{c|}{\textbf{KM}} &
\multicolumn{2}{c|}{\textbf{LO}} &
\multicolumn{2}{c|}{\textbf{MS}} &
\multicolumn{2}{c|}{\textbf{MY}} &
\multicolumn{2}{c|}{\textbf{PT}} &
\multicolumn{2}{c|}{\textbf{TH}} &
\multicolumn{2}{c|}{\textbf{VI}} &
\multicolumn{2}{c|}{\textbf{ZH}} &
\multicolumn{2}{c}{\textbf{Avg.}} \\
\cline{3-26}
& &
NED & TEDS &
NED & TEDS &
NED & TEDS &
NED & TEDS &
NED & TEDS &
NED & TEDS &
NED & TEDS &
NED & TEDS &
NED & TEDS &
NED & TEDS &
NED & TEDS &
NED & TEDS  \\
\hline
\multirow{5}{*}{\textbf{Pipeline Models}}
& \textbf{MinerU2.5}  & 0.071 & 0.913 & 0.061 & 0.918 & 0.051 & 0.932 & 0.54  & 0.394 & 0.648 & 0.366 & 0.079 & 0.908 & 0.308 & 0.666 & 0.059 & 0.922 & 0.416 & 0.515 & 0.109 & 0.872 & 0.055 & 0.929 & 0.128 & 0.851 \\
& \textbf{Dolphin-1.5}& 0.13  & 0.816 & 0.139 & 0.798 & 0.141 & 0.803 & 0.767 & 0.196 & 0.866 & 0.12  & 0.156 & 0.796 & 0.476 & 0.467 & 0.127 & 0.82  & 0.656 & 0.305 & 0.143 & 0.821 & 0.241 & 0.71  & 0.23  & 0.72  \\
& \textbf{MonkeyOCR-pro-1.2B} & 0.083 & 0.896 & 0.119 & 0.863 & 0.106 & 0.874 & 0.71  & 0.254 & 0.764 & 0.281 & 0.14  & 0.836 & 0.468 & 0.524 & 0.13  & 0.856 & 0.511 & 0.52  & 0.188 & 0.808 & 0.128 & 0.835 & 0.196 & 0.789 \\
& \textbf{MonkeyOCR-pro-3B}   & 0.095 & 0.895 & 0.132 & 0.855 & 0.094 & 0.893 & 0.676 & 0.272 & 0.763 & 0.237 & 0.133 & 0.842 & 0.431 & 0.553 & 0.121 & 0.851 & 0.541 & 0.432 & 0.18  & 0.806 & 0.093 & 0.886 & 0.189 & 0.792 \\
& \textbf{PaddleOCR-VL}       & 0.059 & 0.924 & 0.076 & 0.911 & 0.056 & 0.922 & 0.69  & 0.232 & 0.671 & 0.321 & 0.083 & 0.899 & 0.413 & 0.504 & 0.064 & 0.914 & 0.132 & 0.83  & 0.059 & 0.927 & 0.059 & 0.924 & 0.116 & 0.86 \\
\hline
\multirow{3}{*}{\textbf{Expert Models}}
& \textbf{POINTS-Reader} & 0.076 & 0.905 & 0.073 & 0.915 & 0.061 & 0.921 & 0.909 & 0.09  & 0.863 & 0.15  & 0.078 & 0.914 & 0.737 & 0.245 & 0.09  & 0.887 & 0.79  & 0.204 & 0.154 & 0.837 & 0.056 & 0.932 & 0.201 & 0.785 \\
& \textbf{DeepSeek-OCR}  & 0.192 & 0.778 & 0.113 & 0.862 & 0.092 & 0.881 & 0.911 & 0.079 & 0.489 & 0.54  & 0.108 & 0.874 & 0.437 & 0.546 & 0.12  & 0.842 & 0.386 & 0.606 & 0.221 & 0.763 & 0.106 & 0.866 & 0.198 & 0.779 \\
& \textbf{dots.ocr}      & 0.126 & 0.852 & 0.09  & 0.874 & 0.085 & 0.884 & 0.5   & 0.443 & 0.622 & 0.364 & 0.113 & 0.874 & 0.397 & 0.557 & 0.112 & 0.857 & 0.22  & 0.732 & 0.136 & 0.851 & 0.078 & 0.908 & 0.151 & 0.823 \\
\hline
\multirow{5}{*}{\textbf{General Models}}
 & \textbf{InternVL3.5-38B}   & 0.612 & 0.626 & 0.638 & 0.53  & 0.688 & 0.442 & 0.847 & 0.121 & 0.852 & 0.121 & 0.666 & 0.473 & 0.68  & 0.387 & 0.734 & 0.372 & 0.832 & 0.122 & 0.755 & 0.307 & 0.71  & 0.389 & 0.703 & 0.41  \\
& \textbf{Qwen2.5-VL-72B-Instruct} & 0.167 & 0.873 & 0.097 & 0.871 & 0.099 & 0.897 & 0.816 & 0.186 & 0.634 & 0.413 & 0.095 & 0.898 & 0.52  & 0.493 & 0.133 & 0.89  & 0.168 & 0.856 & 0.12  & 0.905 & 0.152 & 0.872 & 0.178 & 0.836 \\
& \textbf{Qwen3-VL-32B-Instruct}   & 0.112 & 0.895 & 0.085 & 0.9   & 0.078 & 0.912 & 0.76  & 0.253 & 0.511 & 0.586 & 0.081 & 0.911 & 0.39  & 0.599 & 0.097 & 0.905 & 0.131 & 0.87  & 0.086 & 0.921 & 0.093 & 0.911 & 0.136 & 0.864 \\
& \textbf{GPT4o}  & 0.199 & 0.761 & 0.125 & 0.85  & 0.172 & 0.77  & 0.534 & 0.358 & 0.577 & 0.376 & 0.176 & 0.823 & 0.332 & 0.57  & 0.311 & 0.65  & 0.271 & 0.667 & 0.248 & 0.744 & 0.275 & 0.603 & 0.248 & 0.702 \\
& \textbf{Gemini2.5-Pro}   & 0.121 & 0.885 & 0.072 & 0.93  & 0.072 & 0.933 & 0.404 & 0.552 & 0.37  & 0.567 & 0.154 & 0.892 & 0.238 & 0.751 & 0.086 & 0.872 & 0.12  & 0.815 & 0.066 & 0.941 & 0.073 & 0.922 & 0.12  & 0.872 \\
\hline
\multicolumn{2}{c|}{\textbf{Avg.}} & 0.157 & 0.848 & 0.140 & 0.852 & 0.138 & 0.851 & 0.697 & 0.264 & 0.664 & 0.342 & 0.159 & 0.842 & 0.448 & 0.528 & 0.168 & 0.818 & 0.398 & 0.575 & 0.190 & 0.808 & 0.163 & 0.822 & 0.215 & 0.776 \\
\hline
\end{tabular}}
\caption{\normalfont Table-region performance (TEDS↑ / NED↓) across 11 languages.}
\label{tab:tables_big}
\end{table*}

\subsubsection{Formulas}
Table \ref{tab:formula_big} shows that formula parsing performance is relatively consistent across model families, with moderate variation between systems. Among pipeline models, PaddleOCR-VL achieves the strongest results (NED 0.230, BLEU 0.586), outperforming Dolphin-1.5 and MinerU2.5. Expert models exhibit comparable performance—dots.ocr and DeepSeek-OCR reach BLEU scores around 0.56–0.58 with NED near 0.27–0.29. General models show the widest spread: Qwen2.5-VL-72B and Qwen3-VL-32B deliver the best overall formula recognition (NED 0.223/0.212, BLEU 0.626/0.633), whereas InternVL3.5-38B lags notably behind. Overall, formula reconstruction remains challenging but stable across most baselines, with top-performing general models showing clear advantages in both structural and token-level accuracy.


\begin{table}[t] %
\setlength{\tabcolsep}{2pt} %
\renewcommand{\arraystretch}{1.15} \resizebox{0.48\textwidth}{!}{ \begin{tabular}{l|l|c|c} \hline \multicolumn{1}{l|}{\textbf{Method Type}} & \multicolumn{1}{l|}{\textbf{Methods}} & \multicolumn{1}{c|}{\textbf{NED}} & \multicolumn{1}{c}{\textbf{BLEU}} \\ \hline \multirow{5}{*}{\textbf{Pipeline Models}} & \textbf{MinerU2.5} & 0.314 & 0.551 \\ & \textbf{Dolphin-1.5} & 0.332 & 0.491 \\ & \textbf{MonkeyOCR-pro-1.2B}& 0.276 & 0.538 \\ & \textbf{MonkeyOCR-pro-3B} & 0.277 & 0.539 \\ & \textbf{PaddleOCR-VL} & 0.230 & 0.586 \\ \hline \multirow{3}{*}{\textbf{Expert Models}} & \textbf{POINTS-Reader} & 0.285 & 0.560\\ & \textbf{DeepSeek-OCR} & 0.291 & 0.581 \\ & \textbf{dots.ocr}& 0.266 & 0.567 \\ \hline \multirow{5}{*}{\textbf{General Models}} & \textbf{InternVL3.5-38B} & 0.521 & 0.329 \\ & \textbf{Qwen2.5-VL-72B-Instruct} & 0.223 & 0.626 \\ & \textbf{Qwen3-VL-32B-Instruct} & 0.212 & 0.633 \\ & \textbf{GPT4o} & 0.379 & 0.451 \\ & \textbf{Gemini2.5-Pro} & 0.267 & 0.578 \\ \hline \multicolumn{2}{c|}{\textbf{Avg.}} & 0.298 & 0.541 \\ \hline \end{tabular} } \caption{\normalfont Formula-region performance (NED↓ / BLEU↑) across 11 languages.} \label{tab:formula_big} \end{table}

\subsubsection{Reading Order}
Table \ref{tab:reading_order} and Table \ref{tab:reading_order_page_type} indicate that most models handle reading-order reconstruction reliably across high-resource scripts and standard page types. Languages such as EN, FIL, ID, and ZH typically show low NED—often around 0.07–0.15—demonstrating that sequential flow can be recovered accurately when script and document conventions are well supported. Larger errors are concentrated in a few low-resource languages, especially Khmer and Lao, where NED may rise to 0.45–0.65 for several baselines. A similar trend appears across page types: single-column formats such as books, exam papers, and slides are consistently easier, while multi-column newspapers and magazines introduce moderate increases in error. Overall, reading-order performance is strong for most languages and page types, with difficulty mainly arising in scripts and structures that deviate from common training distributions.

\begin{table*}[t]
\centering
\setlength{\tabcolsep}{6pt}
\renewcommand{\arraystretch}{1.25}
\label{tab:reading_order}
\resizebox{\textwidth}{!}{
\begin{tabular}{l|l|ccccccccccc|c}
\hline
\multicolumn{1}{c|}{\textbf{Method Type}}  & \multicolumn{1}{c|}{\textbf{Methods}} & \multicolumn{1}{c}{\textbf{EN}} & \multicolumn{1}{c}{\textbf{FIL}} & \multicolumn{1}{c}{\textbf{ID}} & \multicolumn{1}{c}{\textbf{KM}} & \multicolumn{1}{c}{\textbf{LO}} & \multicolumn{1}{c}{\textbf{MS}} & \multicolumn{1}{c}{\textbf{MY}} & \multicolumn{1}{c}{\textbf{PT}} & \multicolumn{1}{c}{\textbf{TH}} & \multicolumn{1}{c}{\textbf{VI}} & \multicolumn{1}{c|}{\textbf{ZH}} & \multicolumn{1}{c}{\textbf{Avg.}} \\
\hline
\multirow{5}{*}{\textbf{Pipeline Models}} & \textbf{MinerU2.5}  & 0.146  & 0.072   & 0.136  & 0.571  & 0.566  & 0.122  & 0.32   & 0.133  & 0.638  & 0.197  & 0.208  & 0.27     \\
& \textbf{Dolphin-1.5}& 0.124  & 0.063   & 0.106  & 0.337  & 0.409  & 0.105  & 0.29   & 0.096  & 0.409  & 0.097  & 0.119  & 0.188    \\
& \textbf{MonkeyOCR-pro-1.2B}  & 0.161  & 0.092   & 0.15   & 0.309  & 0.342  & 0.139  & 0.28   & 0.151  & 0.447  & 0.461  & 0.151  & 0.247    \\
& \textbf{MonkeyOCR-pro-3B}    & 0.161  & 0.087   & 0.123  & 0.331  & 0.327  & 0.11   & 0.255  & 0.135  & 0.494  & 0.271  & 0.15   & 0.221    \\
& \textbf{PaddleOCR-VL}& 0.118  & 0.074   & 0.087  & 0.245  & 0.309  & 0.085  & 0.213  & 0.097  & 0.071  & 0.114  & 0.102  & 0.13     \\
\hline
\multirow{3}{*}{\textbf{Expert Models}}   & \textbf{POINTS-Reader}       & 0.118  & 0.044   & 0.092  & 0.65   & 0.601  & 0.088  & 0.503  & 0.111  & 0.552  & 0.164  & 0.14   & 0.265    \\
& \textbf{DeepSeek-OCR}& 0.169  & 0.069   & 0.117  & 0.452  & 0.105  & 0.121  & 0.251  & 0.141  & 0.222  & 0.19   & 0.131  & 0.182    \\
& \textbf{dots.ocr}   & 0.142  & 0.072   & 0.132  & 0.099  & 0.201  & 0.115  & 0.094  & 0.127  & 0.086  & 0.142  & 0.176  & 0.122    \\
\hline
\multirow{5}{*}{\textbf{General Models}}  & \textbf{InternVL3.5-38B}       & 0.304  & 0.299   & 0.395  & 0.393  & 0.435  & 0.362  & 0.291  & 0.443  & 0.59   & 0.49   & 0.607  & 0.411    \\
& \textbf{Qwen2.5-VL-72B-Instruct}     & 0.189  & 0.086   & 0.139  & 0.61   & 0.306  & 0.121  & 0.366  & 0.133  & 0.08   & 0.12   & 0.187  & 0.204    \\
& \textbf{Qwen3-VL-32B-Instruct}       & 0.142  & 0.064   & 0.12   & 0.497  & 0.203  & 0.112  & 0.302  & 0.107  & 0.075  & 0.109  & 0.126  & 0.165    \\
& \textbf{GPT4o}      & 0.19   & 0.076   & 0.165  & 0.5    & 0.455  & 0.144  & 0.239  & 0.161  & 0.151  & 0.185  & 0.4    & 0.217    \\
& \textbf{Gemini2.5-Pro}       & 0.183  & 0.092   & 0.165  & 0.117  & 0.064  & 0.133  & 0.115  & 0.107  & 0.089  & 0.128  & 0.202  & 0.126    \\
\hline
\multicolumn{2}{c|}{\textbf{Avg.}}      & 0.165  & 0.092   & 0.148  & 0.393  & 0.333  & 0.135  & 0.271  & 0.149  & 0.300  & 0.205  & 0.208  & 0.211  \\
\hline
\end{tabular}
}
\caption{Reading-order NED by language.}
\label{tab:reading_order}
\end{table*}

\begin{table*}[t]
\centering
\setlength{\tabcolsep}{6pt}
\renewcommand{\arraystretch}{1.25}
\label{tab:reading_order_page_type}
\resizebox{\textwidth}{!}{
\begin{tabular}{l|l|ccccccccc|c}
\hline
\multicolumn{1}{c|}{\textbf{Method Type}}  & \multicolumn{1}{c|}{\textbf{Methods}} & \multicolumn{1}{c}{\textbf{Academic paper}} & \multicolumn{1}{c}{\textbf{Book}} & \multicolumn{1}{c}{\textbf{Exam paper}} & \multicolumn{1}{c}{\textbf{Magazine}} & \multicolumn{1}{c}{\textbf{Newspaper}} & \multicolumn{1}{c}{\textbf{Note}} & \multicolumn{1}{c}{\textbf{Research report}} & \multicolumn{1}{c}{\textbf{Slide}} & \multicolumn{1}{c|}{\textbf{Textbook}} & \textbf{Avg.} \\
\hline
\multirow{5}{*}{\textbf{Pipeline Models}} & \textbf{MinerU2.5}   & 0.290       & 0.306     & 0.268   & 0.235 & 0.233  & 0.218     & 0.344& 0.205      & 0.307 & 0.267 \\
  & \textbf{Dolphin-1.5} & 0.145       & 0.179     & 0.231   & 0.187 & 0.199  & 0.158     & 0.203& 0.175      & 0.221 & 0.189 \\
  & \textbf{MonkeyOCR-pro-1.2B}  & 0.208       & 0.233     & 0.273   & 0.277 & 0.336  & 0.198     & 0.271& 0.182      & 0.282 & 0.251 \\
  & \textbf{MonkeyOCR-pro-3B}    & 0.173       & 0.204     & 0.235   & 0.248 & 0.303  & 0.176     & 0.248& 0.171      & 0.265 & 0.225 \\
  & \textbf{PaddleOCR-VL}& 0.076       & 0.118     & 0.149   & 0.157 & 0.187  & 0.096     & 0.145& 0.105      & 0.167 & 0.133 \\
\hline
\multirow{3}{*}{\textbf{Expert Models}}   & \textbf{POINTS-Reader}       & 0.233       & 0.304     & 0.237   & 0.233 & 0.345  & 0.225     & 0.346& 0.163      & 0.302 & 0.265 \\
  & \textbf{DeepSeek-OCR}& 0.155       & 0.189     & 0.182   & 0.212 & 0.285  & 0.127     & 0.173& 0.144      & 0.211 & 0.186 \\
  & \textbf{dots.ocr}    & 0.102       & 0.095     & 0.109   & 0.198 & 0.230  & 0.050     & 0.101& 0.110      & 0.154 & 0.127 \\
\hline
\multirow{5}{*}{\textbf{General Models}}  & \textbf{InternVL3.5-38B}     & 0.497       & 0.470     & 0.328   & 0.488 & 0.612  & 0.268     & 0.484& 0.179      & 0.421 & 0.416 \\
  & \textbf{Qwen2.5-VL-72B-Instruct}     & 0.189       & 0.202     & 0.190   & 0.227 & 0.289  & 0.151     & 0.222& 0.149      & 0.242 & 0.207 \\
  & \textbf{Qwen3-VL-32B-Instruct}       & 0.124       & 0.151     & 0.189   & 0.174 & 0.239  & 0.123     & 0.185& 0.128      & 0.200 & 0.168 \\
  & \textbf{GPT4o}       & 0.199       & 0.232     & 0.191   & 0.264 & 0.368  & 0.136     & 0.263& 0.133      & 0.227 & 0.223 \\
  & \textbf{Gemini2.5-Pro}       & 0.094       & 0.095     & 0.151   & 0.166 & 0.222  & 0.071     & 0.091& 0.121      & 0.174 & 0.132 \\
\hline
\multicolumn{2}{c|}{\textbf{Avg.}}& 0.191       & 0.214     & 0.210   & 0.236 & 0.296  & 0.154     & 0.237& 0.151      & 0.244 & 0.215 \\
\hline
\end{tabular}
}
\caption{Reading-order NED by page type.}
\label{tab:reading_order_page_type}
\end{table*}

\subsection{TEC-VQA Detailed Breakdown}




To highlight the impact of language-resource disparity, we group the 11 languages into high-resource and low-resource sets, following the discussion in the main text. High-resource languages include, for example, English, Chinese, Indonesian, and Malay, which have stronger OCR support and richer pre-training corpora. Low-resource languages include Lao, Khmer, and Burmese, among others.

Table~\ref{tab:tec_resource} summarizes the group-level average TEC-VQA accuracy for these two sets. The average accuracies of MLLMs across the four languages in the high-resource group rank 1st, 2nd, 4th, and 5th among all languages, and their average accuracy (45.78\%) is 2.5 times that of the low-resource group (17.45\%). This reflects a severe lack of attention to low-resource languages among mainstream MLLMs, and a deficiency in multi-scenario recognition and comprehension capabilities for the text of low-resource languages.

\begin{table}[t]
\centering
\caption{TEC-VQA accuracy for high-resource vs.\ low-resource language groups.}
\label{tab:tec_resource}
\resizebox{0.48\textwidth}{!}{
\begin{tabular}{lcc}
\toprule
Language group & Avg. accuracy (\%) &  Languages \\
\midrule
High-resource & 45.78 & EN, ZH, ID, MS\\
\midrule
\multirow{2}{*}{Low-resource}  & \multirow{2}{*}{17.45} & LO, KM, MY, FIL\\
& & PT, TH, VI \\

\bottomrule
\end{tabular}
}
\end{table}

\section{Experimental Details}
This section describes evaluation metrics, baseline configurations, and implementation details used in our experiments.

\subsection{Baseline Configurations}

\subsubsection{Document parsing baselines}
We evaluate three categories of document parsing baselines—pipeline models, expert models, and general models—under a unified protocol. Pipeline systems (MinerU2.5, Dolphin-1.5, MonkeyOCR-pro-1.2B/3B, PaddleOCR-VL) are run with their official checkpoints and default configurations. Expert models (POINTS-Reader, DeepSeek-OCR, dots.ocr) follow their official inference settings, using deterministic decoding when generative components are involved. General multimodal models (InternVL3.5-38B, Qwen2.5-VL-72B-Instruct, Qwen3-VL-32B-Instruct, GPT4o, Gemini2.5-Pro) also adopt deterministic, non-sampling inference (do\_sample=False, temperature~=~0.0). Maximum generation length is fixed per model---e.g., 32k tokens for Qwen2.5-VL-72B and 4096 tokens for InternVL---while closed-source models are evaluated with their default decoding settings.

\subsubsection{TEC-VQA baselines}

For TEC-VQA, we compare both non-generative baselines (for example, classification or retrieval-based models) and generative MLLM-based baselines.

\paragraph{Prompt template.}
All MLLM-based TEC-VQA baselines share a unified prompt template. We use a short system instruction plus an image-conditioned user query. The core user-side template is:

\begin{quote}
\small
\textit{You are a helpful AI assistant that answers questions based on the given image. Please follow these guidelines:\\
1. Answer the question directly and accurately based on what you see in the image, without adding any extra information or explanations. \\
2. If the question is in a specific language, answer in the same language. \\
3. Keep your answer concise and relevant to the question. \\
4. If you cannot find the answer in the image, respond with "No relevant information can be found from the image" or equivalent in the question's language. \\
5. Do not make assumptions or provide information not visible in the image.}\\[0.2em]
\textit{Question (in \{LANG\}): \{QUESTION\}}\\
\textit{Answer (in \{LANG\}, be concise):}
\end{quote}

Here \{LANG\} is replaced by the target language name (for example, “English”, “Vietnamese”), and \{QUESTION\} is the TEC-VQA question text. The image is passed as the visual input channel supported by each MLLM. An illustrative example of the full prompt, including a pseudo document image and question, is shown in Figure~\ref{fig:tec_vqa_prompt_example}.

\begin{figure}[t]
\centering
\caption{Example prompt used for Multimodal Large Language Model (MLLM) TEC-VQA baselines. The document image and question are replaced with actual samples at inference time. (Pseudo example; not from the released dataset.)}
\label{fig:tec_vqa_prompt_example}
\end{figure}

For non-generative baselines that directly classify or retrieve answers from a candidate set, we do not use natural-language prompts and instead follow the original model configuration.

\paragraph{Sampling and decoding settings.}
All TEC-VQA baselines use a unified deterministic setup: no sampling (do\_sample=False, temperature=0), and model-specific maximum output length (e.g., 32k for Qwen2.5-VL-72B, 4k for InternVL3.5-38B). Closed-source models run with their official default decoding settings.

\section{Annotation and Verification Protocols}
In Section~3.2.2 and Section~4 of the main text, we described a hybrid pipeline that combines automatic processing with native-speaker verification for both Document Parsing and TEC-VQA. This part focuses on the four-step human verification protocol used to refine TEC-VQA question–answer (QA) pairs.

\subsection{TEC-VQA Human Verification Checklist}

After automatic generation, every TEC-VQA QA pair is passed through a four-step human verification checklist, which corresponds to the four steps described in Section~3.2.2 of the main paper:

\begin{enumerate}
    \item \textbf{Answerability and usefulness.}
    Annotators first decide whether the question is answerable using only the visible content in the image. Questions that require external knowledge, depend on hidden context, or are overly trivial (e.g., reading a single obvious word with no value for evaluation) are removed.

    \item \textbf{Question clarity and answer normalization.}
    For retained QA pairs, annotators refine the wording of the question to be clear and unambiguous, and normalize the answer format. This includes standardizing number formats (decimal points, thousands separators), making units explicit where needed, and using consistent conventions for dates and currencies across the dataset.

    \item \textbf{Strict alignment with visible text.}
    Annotators then check that all entities appearing in the question and answer (names, numbers, dates, monetary amounts) are present in the visible text of the image and that the answer can be derived solely from that text (possibly with simple reasoning or calculation). If the MLLM misreads a value, hallucinates content that does not exist in the image, or mixes languages, annotators correct the QA pair or discard it if no safe fix exists.

    \item \textbf{Capability labeling.}
    Finally, annotators assign capability labels to each question from the following set: text recognition, numerical calculation, comparative analysis, logical reasoning, and spatial understanding. Multi-label annotations are allowed for questions that require several skills simultaneously (e.g., reading multiple prices and comparing the total).
\end{enumerate}

Table~\ref{tab:b1_qa_evolution} provides an illustrative example of how one QA pair evolves from the original MLLM output to the final, verified version under this checklist.

\begin{table*}[t]
\centering
\caption{Illustrative TEC-VQA example showing one question–answer pair at three stages: original MLLM generation, intermediate human checking, and the final version that satisfies all checklist items. The example is based on a receipt-style image; text here is illustrative.}
\label{tab:b1_qa_evolution}
\resizebox{0.98\textwidth}{!}{%
\begin{tabular}{p{0.16\textwidth}p{0.22\textwidth}p{0.06\textwidth}p{0.50\textwidth}}
\hline
Stage & Question & Answer & Issues / actions \\
\hline
Original generation &
What is the total? &
12.0 &
Question is underspecified (no currency or context); answer format does not match the receipt (“12.80 SGD” printed on the image). \\[0.4em]
\hline
Intermediate check &
What is the total price on the receipt (in dollars)? &
12.50 &
Question is clearer, but the amount is misread (“12.80” on the image). Fails the strict alignment check; needs correction. \\[0.4em]
\hline
Final version &
What is the total amount on the receipt (in SGD)? &
12.80 &
Question specifies the context and currency; answer exactly matches the visible text. Capability labels: text recognition + numerical calculation. \\
\hline
\end{tabular}%
}
\end{table*}

\paragraph{Operational definitions of capability labels.}
For consistency, annotators use short operational rules when assigning capability labels:

\begin{itemize}
    \item \textbf{Text recognition:} the answer is obtained by directly reading and copying a span of visible text (e.g., a store name, product name, or single field).
    \item \textbf{Numerical calculation:} the answer requires explicit arithmetic (addition, subtraction, multiplication, division) or unit conversion using numbers in the image.
    \item \textbf{Comparative analysis:} the answer requires comparing multiple values or text snippets (e.g., choosing the largest price or earliest date).
    \item \textbf{Logical reasoning:} the answer requires combining multiple pieces of information with a condition or rule (e.g., applying a discount rule or free-item policy).
    \item \textbf{Spatial understanding:} the answer depends on layout or positional relations in the image (e.g., “the item listed at the top of the table”).
\end{itemize}

\paragraph{Document Parsing verification.}
Document Parsing labels use the same tool and logging infrastructure but a simpler review protocol: annotators correct semantic categories, and reading order for each page, and a subset of pages is double-checked by a second annotator. Pages containing privacy-sensitive or politically sensitive content are removed before release.

\section{Limitations and License}

\subsection{Limitations}
SEA-Vision is a step toward comprehensive multilingual document and scene text understanding, but it still has several limitations.

First, coverage remains uneven. Some document types and languages are underrepresented, especially highly specialized formats and very low-resource languages, which may bias both training and evaluation. Second, all questions are restricted to those answerable from a single image; cross-document reasoning and the use of external knowledge are not evaluated. Third, despite the multi-stage quality control pipeline, there may be residual noise in OCR text, structural annotations, and question–answer pairs. Fourth, evaluation relies mainly on automatic metrics and does not address interpretability, reasoning transparency, or human preference. Finally, while the benchmark exposes large performance gaps for low-resource languages and complex reasoning skills, it does not yet close them. Extending SEA-Vision and combining it with complementary resources are natural directions for future work.

\subsection{Dataset License and Ethics}
We plan to release SEA-Vision under the Creative Commons Attribution–NonCommercial 4.0 (CC BY-NC 4.0, Creative Commons Attribution–NonCommercial 4.0) license. This allows users to share and adapt the dataset for non-commercial purposes with proper attribution, while restricting direct commercial use.

All documents are collected and processed with privacy and legal considerations in mind. No personally identifiable information is included in the released data; many documents are synthetic or originate from public sources. For real documents (for example, public forms or published materials), we only retain content that is public or permitted for research use. The accompanying documentation will describe data sources, processing steps, and any usage constraints. Users are required not to attempt to reconstruct sensitive information or to deanonymize any content.




\begin{figure*}[!htbp] 
\centering 
\includegraphics[width=0.8\textwidth]{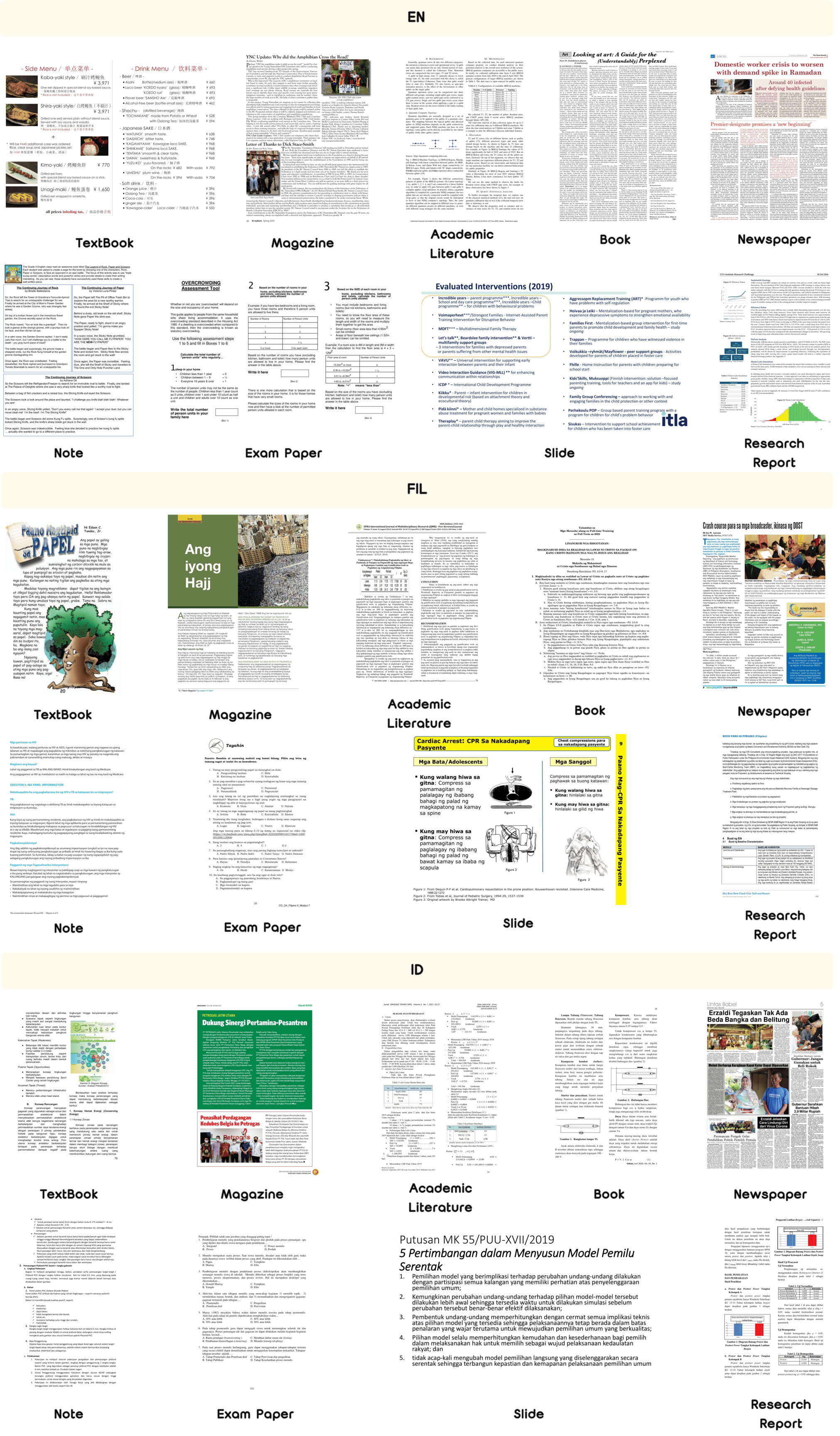} 
\caption{
Representative document page samples for English (EN), Filipino (FIL), and Indonesian (ID) in SEA-Vision.
}
\vspace{-10px}
\label{fig: main} 
\end{figure*}

\begin{figure*}[!htbp] 
\centering 
\includegraphics[width=0.8\textwidth]{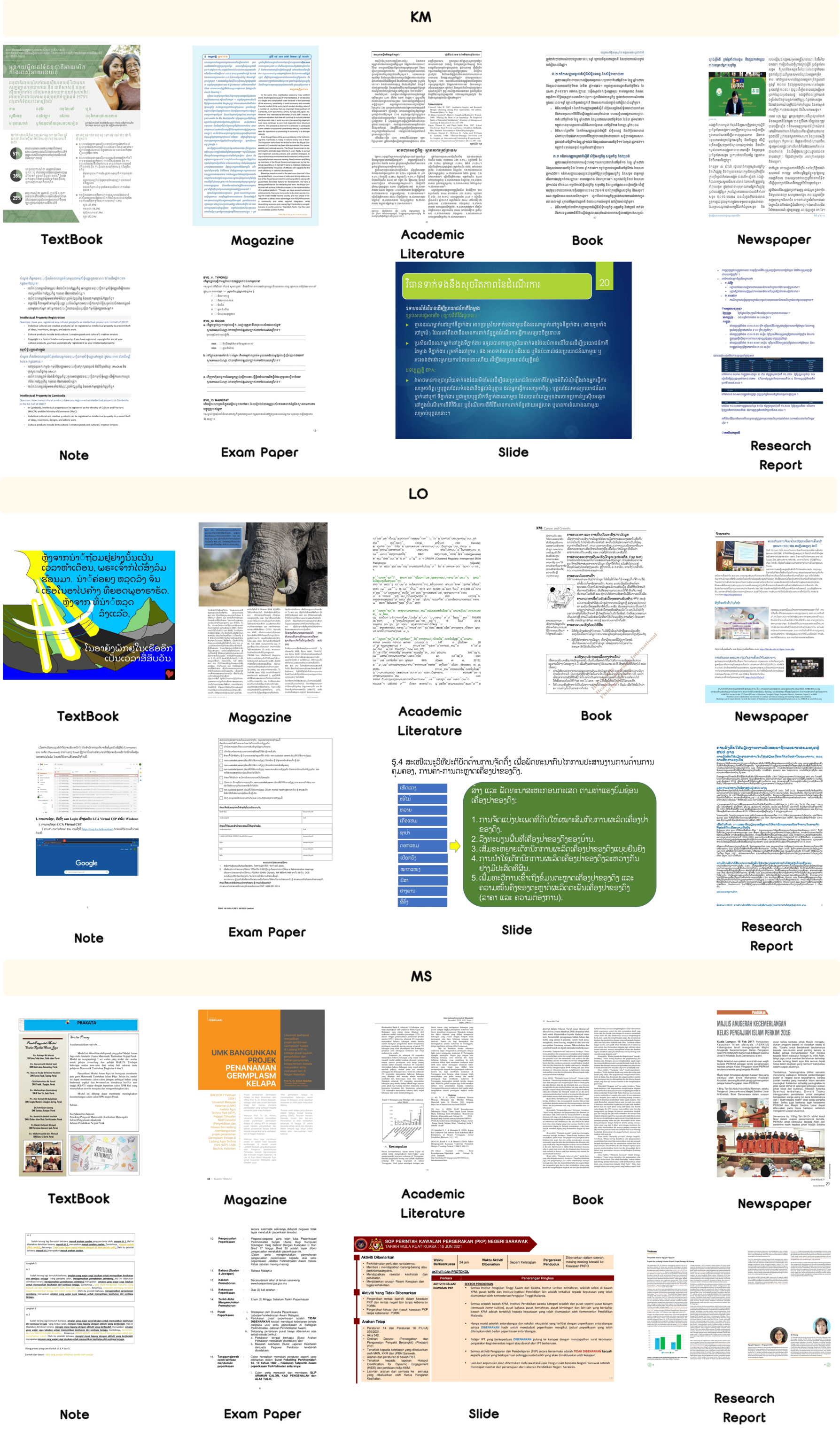} 
\caption{
Representative document page samples for Khmer (KM), Lao (LO), and Malay (MS) in SEA-Vision.
}
\vspace{-10px}
\label{fig: main} 
\end{figure*}

\begin{figure*}[!htbp] 
\centering 
\includegraphics[width=0.75\textwidth]{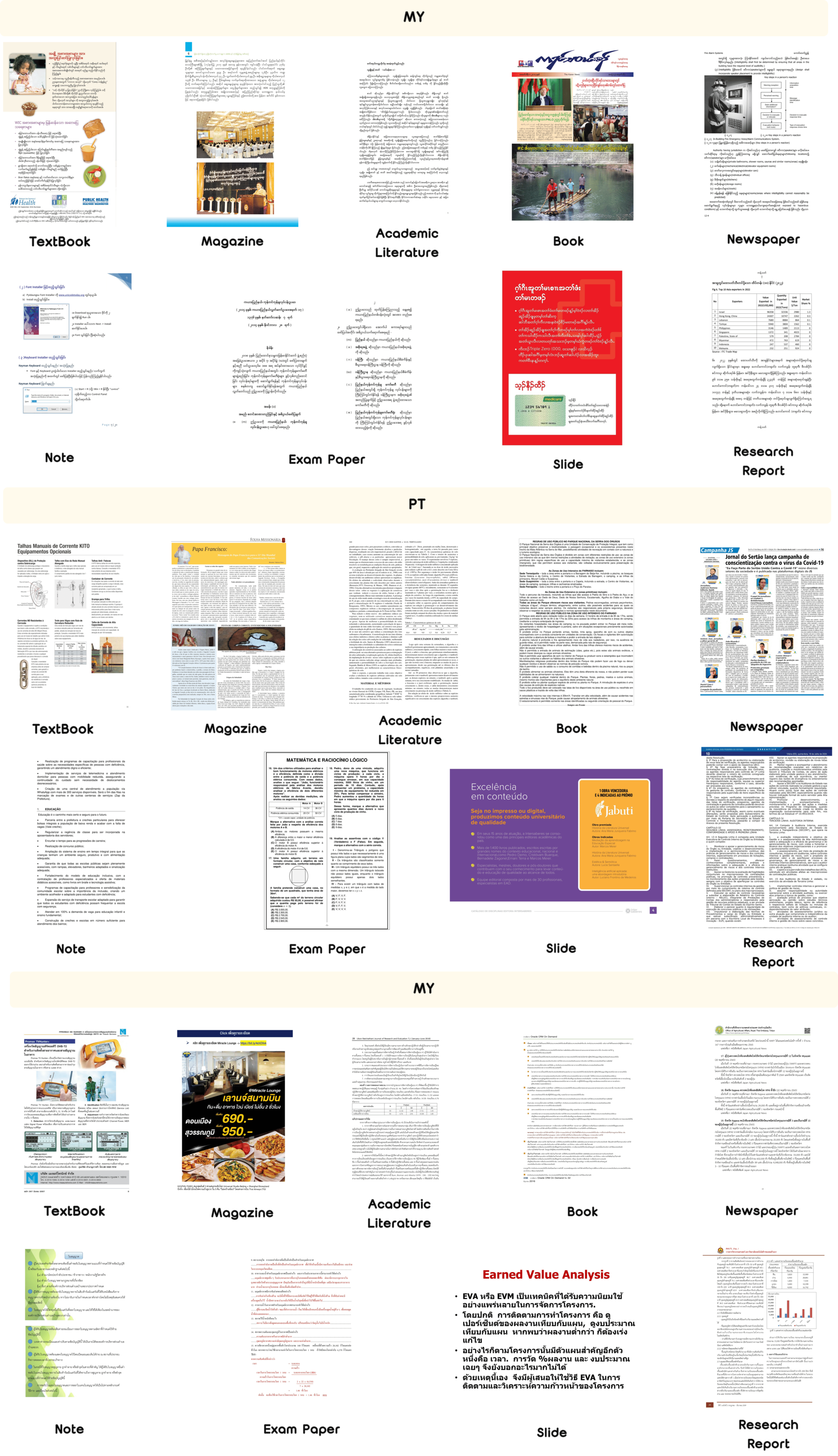} 
\caption{
Representative document page samples for Burmese (MY), Portuguese (PT), and Thai (TH) in SEA-Vision.
}
\vspace{-10px}
\label{fig: main} 
\end{figure*}

\begin{figure*}[!htbp] 
\centering 
\includegraphics[width=0.8\textwidth]{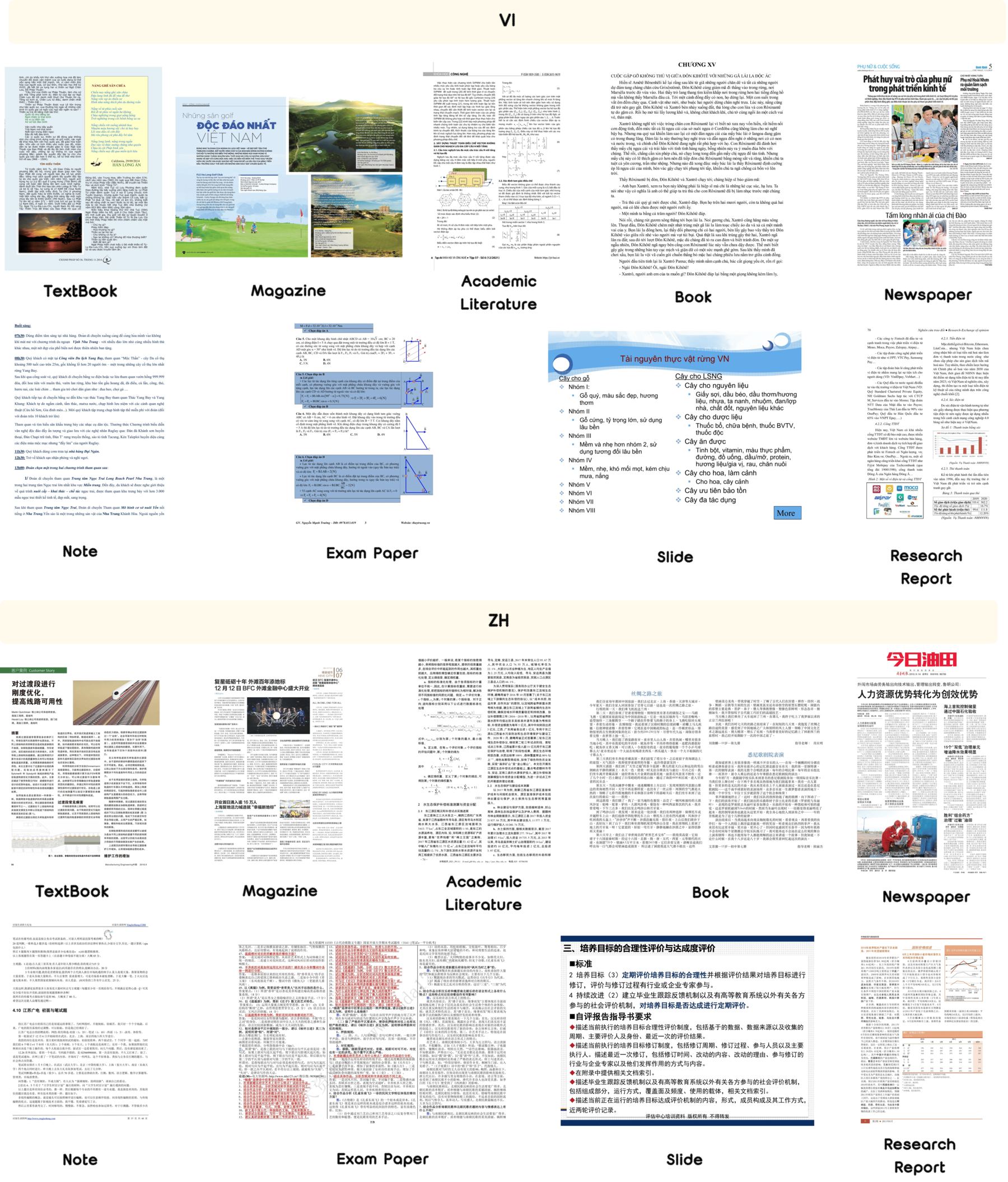} 
\caption{
Representative document page samples for Vietnamese (VI) and Chinese (ZH) in SEA-Vision.
}
\vspace{-10px}
\label{fig: main} 
\end{figure*}

\begin{figure*}[!htbp] 
\centering 
\includegraphics[width=0.75\textwidth]{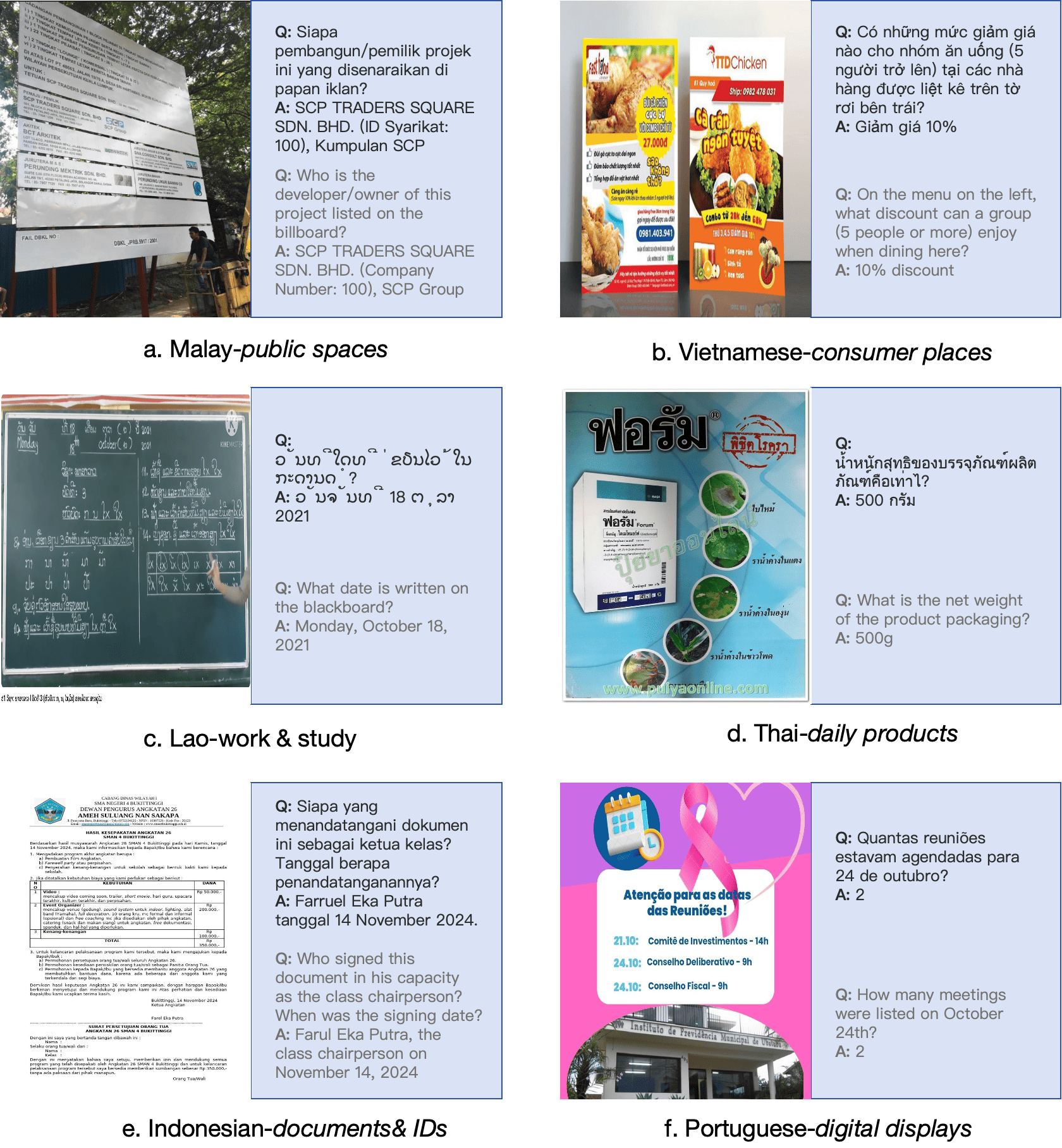} 
\caption{
Representative TEC-VQA examples across multiple Southeast Asian languages.
}
\vspace{-10px}
\label{fig: main} 
\end{figure*}

%% file: main_ref.bib
@String(ICPR = {Int. Conf. Pattern Recog.})

@String(AAAI = {AAAI})

@String(ICPR  = {ICPR})

@article{hurst2024gpt,
  title={Gpt-4o system card},
  author={Hurst, Aaron and Lerer, Adam and Goucher, Adam P and Perelman, Adam and Ramesh, Aditya and Clark, Aidan and Ostrow, AJ and Welihinda, Akila and Hayes, Alan and Radford, Alec and others},
  journal={arXiv preprint arXiv:2410.21276},
  year={2024}
}

@misc{qwen3technicalreport,
      title={Qwen3 Technical Report}, 
      author={Qwen Team},
      year={2025},
      eprint={2505.09388},
      archivePrefix={arXiv},
      primaryClass={cs.CL},
      url={https://arxiv.org/abs/2505.09388}, 
}

@article{wang2025internvl3_5,
  title={Internvl3. 5: Advancing open-source multimodal models in versatility, reasoning, and efficiency},
  author={Wang, Weiyun and Gao, Zhangwei and Gu, Lixin and Pu, Hengjun and Cui, Long and Wei, Xingguang and Liu, Zhaoyang and Jing, Linglin and Ye, Shenglong and Shao, Jie and others},
  journal={arXiv preprint arXiv:2508.18265},
  year={2025}
}

@inproceedings{nayef2019icdar2019,
  title={Icdar2019 robust reading challenge on multi-lingual scene text detection and recognition—rrc-mlt-2019},
  author={Nayef, Nibal and Patel, Yash and Busta, Michal and Chowdhury, Pinaki Nath and Karatzas, Dimosthenis and Khlif, Wafa and Matas, Jiri and Pal, Umapada and Burie, Jean-Christophe and Liu, Cheng-lin and others},
  booktitle={2019 International conference on document analysis and recognition (ICDAR)},
  pages={1582--1587},
  year={2019},
  organization={IEEE}
}

@inproceedings{karatzas2015icdar,
  title={ICDAR 2015 competition on robust reading},
  author={Karatzas, Dimosthenis and Gomez-Bigorda, Lluis and Nicolaou, Anguelos and Ghosh, Suman and Bagdanov, Andrew and Iwamura, Masakazu and Matas, Jiri and Neumann, Lukas and Chandrasekhar, Vijay Ramaseshan and Lu, Shijian and others},
  booktitle={2015 13th international conference on document analysis and recognition (ICDAR)},
  pages={1156--1160},
  year={2015},
  organization={IEEE}
}

@inproceedings{karatzas2013icdar,
  title={ICDAR 2013 robust reading competition},
  author={Karatzas, Dimosthenis and Shafait, Faisal and Uchida, Seiichi and Iwamura, Masakazu and i Bigorda, Lluis Gomez and Mestre, Sergi Robles and Mas, Joan and Mota, David Fernandez and Almazan, Jon Almazan and De Las Heras, Lluis Pere},
  booktitle={2013 12th international conference on document analysis and recognition},
  pages={1484--1493},
  year={2013},
  organization={IEEE}
}

@inproceedings{jiu2025tvqacml,
  title={TVQACML: Benchmarking Text-Centric Visual Question Answering in Multilingual Chinese Minority Languages},
  author={Jiu, Sha and Weng, Yu and Zhu, Mengxiao and Feng, Chong and Liu, Zheng and others},
  booktitle={Proceedings of the 2025 Conference on Empirical Methods in Natural Language Processing},
  pages={13968--13978},
  year={2025}
}

@article{li2020docbank,
  title={Docbank: A benchmark dataset for document layout analysis},
  author={Li, Minghao and Xu, Yiheng and Cui, Lei and Huang, Shaohan and Wei, Furu and Li, Zhoujun and Zhou, Ming},
  journal={arXiv preprint arXiv:2006.01038},
  year={2020}
}

@inproceedings{pfitzmann2022doclaynet,
  title={Doclaynet: A large human-annotated dataset for document-layout segmentation},
  author={Pfitzmann, Birgit and Auer, Christoph and Dolfi, Michele and Nassar, Ahmed S and Staar, Peter},
  booktitle={Proceedings of the 28th ACM SIGKDD conference on knowledge discovery and data mining},
  pages={3743--3751},
  year={2022}
}

@article{wang2024cdm,
  title={Cdm: A reliable metric for fair and accurate formula recognition evaluation},
  author={Wang, Bin and Wu, Fan and Ouyang, Linke and Gu, Zhuangcheng and Zhang, Rui and Xia, Renqiu and Zhang, Bo and He, Conghui},
  journal={arXiv e-prints},
  pages={arXiv--2409},
  year={2024}
}

@inproceedings{papineni2002bleu,
  title={Bleu: a method for automatic evaluation of machine translation},
  author={Papineni, Kishore and Roukos, Salim and Ward, Todd and Zhu, Wei-Jing},
  booktitle={Proceedings of the 40th annual meeting of the Association for Computational Linguistics},
  pages={311--318},
  year={2002}
}

@inproceedings{lcvenshtcin1966binary,
  title={Binary coors capable or ‘correcting deletions, insertions, and reversals},
  author={Lcvenshtcin, VI},
  booktitle={Soviet physics-doklady},
  volume={10},
  number={8},
  year={1966}
}

@inproceedings{zhong2020teds,
  title={Image-based table recognition: data, model, and evaluation},
  author={Zhong, Xu and ShafieiBavani, Elaheh and Jimeno Yepes, Antonio},
  booktitle={European conference on computer vision},
  pages={564--580},
  year={2020},
  organization={Springer}
}

@article{wang2021layoutreader,
  title={Layoutreader: Pre-training of text and layout for reading order detection},
  author={Wang, Zilong and Xu, Yiheng and Cui, Lei and Shang, Jingbo and Wei, Furu},
  journal={arXiv preprint arXiv:2108.11591},
  year={2021}
}

@inproceedings{gu2022xylayoutlm,
  title={Xylayoutlm: Towards layout-aware multimodal networks for visually-rich document understanding},
  author={Gu, Zhangxuan and Meng, Changhua and Wang, Ke and Lan, Jun and Wang, Weiqiang and Gu, Ming and Zhang, Liqing},
  booktitle={Proceedings of the IEEE/CVF conference on computer vision and pattern recognition},
  pages={4583--4592},
  year={2022}
}

@inproceedings{biten2019scenetextvqa,
  title={Scene text visual question answering},
  author={Biten, Ali Furkan and Tito, Ruben and Mafla, Andres and Gomez, Lluis and Rusinol, Mar{\c{c}}al and Valveny, Ernest and Jawahar, CV and Karatzas, Dimosthenis},
  booktitle={Proceedings of the IEEE/CVF international conference on computer vision},
  pages={4291--4301},
  year={2019}
}

@article{liu2024ocrbench,
  title={Ocrbench: on the hidden mystery of ocr in large multimodal models},
  author={Liu, Yuliang and Li, Zhang and Huang, Mingxin and Yang, Biao and Yu, Wenwen and Li, Chunyuan and Yin, Xu-Cheng and Liu, Cheng-Lin and Jin, Lianwen and Bai, Xiang},
  journal={Science China Information Sciences},
  volume={67},
  number={12},
  pages={220102},
  year={2024},
  publisher={Springer}
}

@article{fu2024ocrbenchv2,
  title={Ocrbench v2: An improved benchmark for evaluating large multimodal models on visual text localization and reasoning},
  author={Fu, Ling and Kuang, Zhebin and Song, Jiajun and Huang, Mingxin and Yang, Biao and Li, Yuzhe and Zhu, Linghao and Luo, Qidi and Wang, Xinyu and Lu, Hao and others},
  journal={arXiv preprint arXiv:2501.00321},
  year={2024}
}

@article{comanici2025gemini,
  title={Gemini 2.5: Pushing the frontier with advanced reasoning, multimodality, long context, and next generation agentic capabilities},
  author={Comanici, Gheorghe and Bieber, Eric and Schaekermann, Mike and Pasupat, Ice and Sachdeva, Noveen and Dhillon, Inderjit and Blistein, Marcel and Ram, Ori and Zhang, Dan and Rosen, Evan and others},
  journal={arXiv preprint arXiv:2507.06261},
  year={2025}
}

@article{zhu2025internvl3,
  title={Internvl3: Exploring advanced training and test-time recipes for open-source multimodal models},
  author={Zhu, Jinguo and Wang, Weiyun and Chen, Zhe and Liu, Zhaoyang and Ye, Shenglong and Gu, Lixin and Tian, Hao and Duan, Yuchen and Su, Weijie and Shao, Jie and others},
  journal={arXiv preprint arXiv:2504.10479},
  year={2025}
}

@article{yu2025minicpm4_5,
  title={Minicpm-v 4.5: Cooking efficient mllms via architecture, data, and training recipe},
  author={Yu, Tianyu and Wang, Zefan and Wang, Chongyi and Huang, Fuwei and Ma, Wenshuo and He, Zhihui and Cai, Tianchi and Chen, Weize and Huang, Yuxiang and Zhao, Yuanqian and others},
  journal={arXiv preprint arXiv:2509.18154},
  year={2025}
}

@article{yao2024minicpm2_6,
  title={Minicpm-v: A gpt-4v level mllm on your phone},
  author={Yao, Yuan and Yu, Tianyu and Zhang, Ao and Wang, Chongyi and Cui, Junbo and Zhu, Hongji and Cai, Tianchi and Li, Haoyu and Zhao, Weilin and He, Zhihui and others},
  journal={arXiv preprint arXiv:2408.01800},
  year={2024}
}

@article{wu2024deepseek,
  title={Deepseek-vl2: Mixture-of-experts vision-language models for advanced multimodal understanding},
  author={Wu, Zhiyu and Chen, Xiaokang and Pan, Zizheng and Liu, Xingchao and Liu, Wen and Dai, Damai and Gao, Huazuo and Ma, Yiyang and Wu, Chengyue and Wang, Bingxuan and others},
  journal={arXiv preprint arXiv:2412.10302},
  year={2024}
}

@inproceedings{brandt2008ovis,
  title={OVIS-2: A robust distributed architecture for scalable RAS},
  author={Brandt, Jim M and Debusschere, Bert J and Gentile, Ann C and Mayo, Jackson R and P{\'e}bay, Philippe P and Thompson, David and Wong, Matthew H},
  booktitle={2008 IEEE International Symposium on Parallel and Distributed Processing},
  pages={1--8},
  year={2008},
  organization={IEEE}
}

@article{an2025llava,
  title={Llava-onevision-1.5: Fully open framework for democratized multimodal training},
  author={An, Xiang and Xie, Yin and Yang, Kaicheng and Zhang, Wenkang and Zhao, Xiuwei and Cheng, Zheng and Wang, Yirui and Xu, Songcen and Chen, Changrui and Wu, Chunsheng and others},
  journal={arXiv preprint arXiv:2509.23661},
  year={2025}
}

@inproceedings{mathew2021docvqa,
  title={Docvqa: A dataset for vqa on document images},
  author={Mathew, Minesh and Karatzas, Dimosthenis and Jawahar, CV},
  booktitle={Proceedings of the IEEE/CVF winter conference on applications of computer vision},
  pages={2200--2209},
  year={2021}
}

@inproceedings{mathew2022infographicvqa,
  title={Infographicvqa},
  author={Mathew, Minesh and Bagal, Viraj and Tito, Rub{\`e}n and Karatzas, Dimosthenis and Valveny, Ernest and Jawahar, CV},
  booktitle={Proceedings of the IEEE/CVF Winter Conference on Applications of Computer Vision},
  pages={1697--1706},
  year={2022}
}

@inproceedings{yang2025cc,
  title={Cc-ocr: A comprehensive and challenging ocr benchmark for evaluating large multimodal models in literacy},
  author={Yang, Zhibo and Tang, Jun and Li, Zhaohai and Wang, Pengfei and Wan, Jianqiang and Zhong, Humen and Liu, Xuejing and Yang, Mingkun and Wang, Peng and Bai, Shuai and others},
  booktitle={Proceedings of the IEEE/CVF International Conference on Computer Vision},
  pages={21744--21754},
  year={2025}
}

@article{zhang2023m3exam,
  title={M3exam: A multilingual, multimodal, multilevel benchmark for examining large language models},
  author={Zhang, Wenxuan and Aljunied, Mahani and Gao, Chang and Chia, Yew Ken and Bing, Lidong},
  journal={Advances in Neural Information Processing Systems},
  volume={36},
  pages={5484--5505},
  year={2023}
}

@inproceedings{singh2019textvqa,
  title={Towards vqa models that can read},
  author={Singh, Amanpreet and Natarajan, Vivek and Shah, Meet and Jiang, Yu and Chen, Xinlei and Batra, Dhruv and Parikh, Devi and Rohrbach, Marcus},
  booktitle={Proceedings of the IEEE/CVF conference on computer vision and pattern recognition},
  pages={8317--8326},
  year={2019}
}

@article{Qwen2.5-VL,
  title={Qwen2.5-VL Technical Report},
  author={Bai, Shuai and Chen, Keqin and Liu, Xuejing and Wang, Jialin and Ge, Wenbin and Song, Sibo and Dang, Kai and Wang, Peng and Wang, Shijie and Tang, Jun and Zhong, Humen and Zhu, Yuanzhi and Yang, Mingkun and Li, Zhaohai and Wan, Jianqiang and Wang, Pengfei and Ding, Wei and Fu, Zheren and Xu, Yiheng and Ye, Jiabo and Zhang, Xi and Xie, Tianbao and Cheng, Zesen and Zhang, Hang and Yang, Zhibo and Xu, Haiyang and Lin, Junyang},
  journal={arXiv preprint arXiv:2502.13923},
  year={2025}
}

@article{Qwen2-VL,
  title={Qwen2-VL: Enhancing Vision-Language Model's Perception of the World at Any Resolution},
  author={Wang, Peng and Bai, Shuai and Tan, Sinan and Wang, Shijie and Fan, Zhihao and Bai, Jinze and Chen, Keqin and Liu, Xuejing and Wang, Jialin and Ge, Wenbin and Fan, Yang and Dang, Kai and Du, Mengfei and Ren, Xuancheng and Men, Rui and Liu, Dayiheng and Zhou, Chang and Zhou, Jingren and Lin, Junyang},
  journal={arXiv preprint arXiv:2409.12191},
  year={2024}
}

@article{Qwen-VL,
  title={Qwen-VL: A Versatile Vision-Language Model for Understanding, Localization, Text Reading, and Beyond},
  author={Bai, Jinze and Bai, Shuai and Yang, Shusheng and Wang, Shijie and Tan, Sinan and Wang, Peng and Lin, Junyang and Zhou, Chang and Zhou, Jingren},
  journal={arXiv preprint arXiv:2308.12966},
  year={2023}
}

@article{feng2024docpedia,
  title={Docpedia: Unleashing the power of large multimodal model in the frequency domain for versatile document understanding},
  author={Feng, Hao and Liu, Qi and Liu, Hao and Tang, Jingqun and Zhou, Wengang and Li, Houqiang and Huang, Can},
  journal={Science China Information Sciences},
  volume={67},
  number={12},
  pages={220106},
  year={2024},
  publisher={Springer}
}

@article{zhao2024harmonizing,
  title={Harmonizing visual text comprehension and generation},
  author={Zhao, Zhen and Tang, Jingqun and Wu, Binghong and Lin, Chunhui and Wei, Shu and Liu, Hao and Tan, Xin and Zhang, Zhizhong and Huang, Can and Xie, Yuan},
  journal={Advances in Neural Information Processing Systems},
  volume={37},
  pages={97499--97522},
  year={2024}
}

@article{liu2024textmonkey,
  title={Textmonkey: An ocr-free large multimodal model for understanding document},
  author={Liu, Yuliang and Yang, Biao and Liu, Qiang and Li, Zhang and Ma, Zhiyin and Zhang, Shuo and Bai, Xiang},
  journal={arXiv preprint arXiv:2403.04473},
  year={2024}
}

@article{feng2023unidoc,
  title={Unidoc: A universal large multimodal model for simultaneous text detection, recognition, spotting and understanding},
  author={Feng, Hao and Wang, Zijian and Tang, Jingqun and Lu, Jinghui and Zhou, Wengang and Li, Houqiang and Huang, Can},
  journal={arXiv preprint arXiv:2308.11592},
  year={2023}
}

@article{zhang2023llavar,
  title={Llavar: Enhanced visual instruction tuning for text-rich image understanding},
  author={Zhang, Yanzhe and Zhang, Ruiyi and Gu, Jiuxiang and Zhou, Yufan and Lipka, Nedim and Yang, Diyi and Sun, Tong},
  journal={arXiv preprint arXiv:2306.17107},
  year={2023}
}

@inproceedings{tang2025mtvqa,
  title={Mtvqa: Benchmarking multilingual text-centric visual question answering},
  author={Tang, Jingqun and Liu, Qi and Ye, Yongjie and Lu, Jinghui and Wei, Shu and Wang, An-Lan and Lin, Chunhui and Feng, Hao and Zhao, Zhen and Wang, Yanjie and others},
  booktitle={Findings of the Association for Computational Linguistics: ACL 2025},
  pages={7748--7763},
  year={2025}
}

@inproceedings{antol2015vqa,
  title={Vqa: Visual question answering},
  author={Antol, Stanislaw and Agrawal, Aishwarya and Lu, Jiasen and Mitchell, Margaret and Batra, Dhruv and Zitnick, C Lawrence and Parikh, Devi},
  booktitle={Proceedings of the IEEE international conference on computer vision},
  pages={2425--2433},
  year={2015}
}

@article{ye2023mplug,
  title={mplug-docowl: Modularized multimodal large language model for document understanding},
  author={Ye, Jiabo and Hu, Anwen and Xu, Haiyang and Ye, Qinghao and Yan, Ming and Dan, Yuhao and Zhao, Chenlin and Xu, Guohai and Li, Chenliang and Tian, Junfeng and others},
  journal={arXiv preprint arXiv:2307.02499},
  year={2023}
}

@article{wang2024mineru,
  title={Mineru: An open-source solution for precise document content extraction},
  author={Wang, Bin and Xu, Chao and Zhao, Xiaomeng and Ouyang, Linke and Wu, Fan and Zhao, Zhiyuan and Xu, Rui and Liu, Kaiwen and Qu, Yuan and Shang, Fukai and others},
  journal={arXiv preprint arXiv:2409.18839},
  year={2024}
}

@inproceedings{ouyang2025omnidocbench,
  title={Omnidocbench: Benchmarking diverse pdf document parsing with comprehensive annotations},
  author={Ouyang, Linke and Qu, Yuan and Zhou, Hongbin and Zhu, Jiawei and Zhang, Rui and Lin, Qunshu and Wang, Bin and Zhao, Zhiyuan and Jiang, Man and Zhao, Xiaomeng and others},
  booktitle={Proceedings of the Computer Vision and Pattern Recognition Conference},
  pages={24838--24848},
  year={2025}
}

@article{niu2025mineru2,
  title={Mineru2. 5: A decoupled vision-language model for efficient high-resolution document parsing},
  author={Niu, Junbo and Liu, Zheng and Gu, Zhuangcheng and Wang, Bin and Ouyang, Linke and Zhao, Zhiyuan and Chu, Tao and He, Tianyao and Wu, Fan and Zhang, Qintong and others},
  journal={arXiv preprint arXiv:2509.22186},
  year={2025}
}

@article{feng2025dolphin,
  title={Dolphin: Document image parsing via heterogeneous anchor prompting},
  author={Feng, Hao and Wei, Shu and Fei, Xiang and Shi, Wei and Han, Yingdong and Liao, Lei and Lu, Jinghui and Wu, Binghong and Liu, Qi and Lin, Chunhui and others},
  journal={arXiv preprint arXiv:2505.14059},
  year={2025}
}

@article{li2025monkeyocr,
  title={MonkeyOCR: Document Parsing with a Structure-Recognition-Relation Triplet Paradigm},
  author={Li, Zhang and Liu, Yuliang and Liu, Qiang and Ma, Zhiyin and Zhang, Ziyang and Zhang, Shuo and Guo, Zidun and Zhang, Jiarui and Wang, Xinyu and Bai, Xiang},
  journal={arXiv preprint arXiv:2506.05218},
  year={2025}
}

@article{wei2025deepseekocr,
  title={DeepSeek-OCR: Contexts Optical Compression},
  author={Wei, Haoran and Sun, Yaofeng and Li, Yukun},
  journal={arXiv preprint arXiv:2510.18234},
  year={2025}
}

@article{cui2025paddleocr,
  title={PaddleOCR-VL: Boosting Multilingual Document Parsing via a 0.9 B Ultra-Compact Vision-Language Model},
  author={Cui, Cheng and Sun, Ting and Liang, Suyin and Gao, Tingquan and Zhang, Zelun and Liu, Jiaxuan and Wang, Xueqing and Zhou, Changda and Liu, Hongen and Lin, Manhui and others},
  journal={arXiv preprint arXiv:2510.14528},
  year={2025}
}

@article{zhao2024doclayout,
  title={Doclayout-yolo: Enhancing document layout analysis through diverse synthetic data and global-to-local adaptive perception},
  author={Zhao, Zhiyuan and Kang, Hengrui and Wang, Bin and He, Conghui},
  journal={arXiv preprint arXiv:2410.12628},
  year={2024}
}

@article{gu2021unidoc,
  title={Unidoc: Unified pretraining framework for document understanding},
  author={Gu, Jiuxiang and Kuen, Jason and Morariu, Vlad I and Zhao, Handong and Jain, Rajiv and Barmpalios, Nikolaos and Nenkova, Ani and Sun, Tong},
  journal={Advances in Neural Information Processing Systems},
  volume={34},
  pages={39--50},
  year={2021}
}

@inproceedings{huang2022swintextspotter,
  title={Swintextspotter: Scene text spotting via better synergy between text detection and text recognition},
  author={Huang, Mingxin and Liu, Yuliang and Peng, Zhenghao and Liu, Chongyu and Lin, Dahua and Zhu, Shenggao and Yuan, Nicholas and Ding, Kai and Jin, Lianwen},
  booktitle={proceedings of the IEEE/CVF conference on computer vision and pattern recognition},
  pages={4593--4603},
  year={2022}
}

@article{li2022ppocrv3,
  title={PP-OCRv3: More attempts for the improvement of ultra lightweight OCR system},
  author={Li, Chenxia and Liu, Weiwei and Guo, Ruoyu and Yin, Xiaoting and Jiang, Kaitao and Du, Yongkun and Du, Yuning and Zhu, Lingfeng and Lai, Baohua and Hu, Xiaoguang and others},
  journal={arXiv preprint arXiv:2206.03001},
  year={2022}
}

@inproceedings{liu2020abcnet,
  title={Abcnet: Real-time scene text spotting with adaptive bezier-curve network},
  author={Liu, Yuliang and Chen, Hao and Shen, Chunhua and He, Tong and Jin, Lianwen and Wang, Liangwei},
  booktitle={proceedings of the IEEE/CVF conference on computer vision and pattern recognition},
  pages={9809--9818},
  year={2020}
}

@inproceedings{smith2009adapting,
  title={Adapting the Tesseract open source OCR engine for multilingual OCR},
  author={Smith, Ray and Antonova, Daria and Lee, Dar-Shyang},
  booktitle={Proceedings of the international workshop on multilingual OCR},
  pages={1--8},
  year={2009}
}

@inproceedings{wang2021pgnet,
  title={Pgnet: Real-time arbitrarily-shaped text spotting with point gathering network},
  author={Wang, Pengfei and Zhang, Chengquan and Qi, Fei and Liu, Shanshan and Zhang, Xiaoqiang and Lyu, Pengyuan and Han, Junyu and Liu, Jingtuo and Ding, Errui and Shi, Guangming},
  booktitle={Proceedings of the AAAI Conference on Artificial Intelligence},
  volume={35},
  number={4},
  pages={2782--2790},
  year={2021}
}

@article{pramanik2020towards,
  title={Towards a multi-modal, multi-task learning based pre-training framework for document representation learning},
  author={Pramanik, Subhojeet and Mujumdar, Shashank and Patel, Hima},
  journal={arXiv preprint arXiv:2009.14457},
  year={2020}
}

@inproceedings{huang2022layoutlmv3,
  title={Layoutlmv3: Pre-training for document ai with unified text and image masking},
  author={Huang, Yupan and Lv, Tengchao and Cui, Lei and Lu, Yutong and Wei, Furu},
  booktitle={Proceedings of the 30th ACM international conference on multimedia},
  pages={4083--4091},
  year={2022}
}

@inproceedings{liu2025points,
  title={POINTS-Reader: Distillation-Free Adaptation of Vision-Language Models for Document Conversion},
  author={Liu, Yuan and Zhao, Zhongyin and Tian, Le and Wang, Haicheng and Ye, Xubing and You, Yangxiu and Yu, Zilin and Wu, Chuhan and Xiao, Zhou and Yu, Yang and others},
  booktitle={Proceedings of the 2025 Conference on Empirical Methods in Natural Language Processing},
  pages={1576--1601},
  year={2025}
}

@software{dots.ocr,
  title        = {dots.ocr: Multilingual Document Layout Parsing in a Single Vision‑Language Model},
  author       = {rednote‑hilab},
  year         = {2025},
  url          = {https://github.com/rednote-hilab/dots.ocr},
  note         = {GitHub repository},
}

@article{wang2024unimernet,
  title={Unimernet: A universal network for real-world mathematical expression recognition},
  author={Wang, Bin and Gu, Zhuangcheng and Liang, Guang and Xu, Chao and Zhang, Bo and Shi, Botian and He, Conghui},
  journal={arXiv preprint arXiv:2404.15254},
  year={2024}
}

@inproceedings{li2020improvingformula,
  title={Improving attention-based handwritten mathematical expression recognition with scale augmentation and drop attention},
  author={Li, Zhe and Jin, Lianwen and Lai, Songxuan and Zhu, Yecheng},
  booktitle={2020 17th International Conference on Frontiers in Handwriting Recognition (ICFHR)},
  pages={175--180},
  year={2020},
  organization={IEEE}
}

@inproceedings{zhang2018multiformula,
  title={Multi-scale attention with dense encoder for handwritten mathematical expression recognition},
  author={Zhang, Jianshu and Du, Jun and Dai, Lirong},
  booktitle={2018 24th international conference on pattern recognition (ICPR)},
  pages={2245--2250},
  year={2018},
  organization={IEEE}
}

@article{huang2020tabtransformer,
  title={Tabtransformer: Tabular data modeling using contextual embeddings},
  author={Huang, Xin and Khetan, Ashish and Cvitkovic, Milan and Karnin, Zohar},
  journal={arXiv preprint arXiv:2012.06678},
  year={2020}
}

@inproceedings{huang2023improvingtable,
  title={Improving table structure recognition with visual-alignment sequential coordinate modeling},
  author={Huang, Yongshuai and Lu, Ning and Chen, Dapeng and Li, Yibo and Xie, Zecheng and Zhu, Shenggao and Gao, Liangcai and Peng, Wei},
  booktitle={Proceedings of the IEEE/CVF Conference on Computer Vision and Pattern Recognition},
  pages={11134--11143},
  year={2023}
}

@article{liu2024focus,
  title={Focus anywhere for fine-grained multi-page document understanding},
  author={Liu, Chenglong and Wei, Haoran and Chen, Jinyue and Kong, Lingyu and Ge, Zheng and Zhu, Zining and Zhao, Liang and Sun, Jianjian and Han, Chunrui and Zhang, Xiangyu},
  journal={arXiv preprint arXiv:2405.14295},
  year={2024}
}
